\documentclass[a4paper]{llncs}

\usepackage[T1]{fontenc}
\usepackage[utf8]{inputenc}
\usepackage{amsmath,amssymb}
\usepackage{times,helvet,courier}
\usepackage{url}
\usepackage{xspace}
\usepackage[vlined,linesnumbered]{algorithm2e}
\usepackage{listings}
\usepackage{tikz}
\usepackage{floatrow}
\usepackage{centernot}
\usepackage{xcolor}
\usetikzlibrary{positioning,calc,fit,shapes}

\SetCommentSty{textnormal}

\newcommand{\Sum}{\ensuremath{\mathit{sum}}\xspace}
\newcommand{\Count}{\ensuremath{\mathit{count}}\xspace}
\newcommand{\Min}{\ensuremath{\mathit{min}}\xspace}
\newcommand{\Max}{\ensuremath{\mathit{max}}\xspace}
\newcommand{\Naf}{\ensuremath{{\sim}}}
\newcommand{\Weight}{\ensuremath{\mathit{weight}}\xspace}
\newcommand{\Pred}[1]{\ensuremath{\mathit{#1}}\xspace}
\newcommand{\Tuple}{\ensuremath{\mathit{tuple}}\xspace}
\newcommand{\Body}{\ensuremath{\mathrm{body}}}
\newcommand{\Head}{\ensuremath{\mathrm{head}}}
\newcommand{\Accu}{\ensuremath{\mathit{accu}}}
\newcommand{\Aggr}{\ensuremath{\mathit{aggr}}}
\newcommand{\Neutral}{\ensuremath{\mathit{neutral}}\xspace}

\newcommand{\DomNew}{\ensuremath{\mathrm{n}}\xspace}
\newcommand{\DomOld}{\ensuremath{\mathrm{o}}\xspace}
\newcommand{\DomAll}{\ensuremath{\mathrm{a}}\xspace}
\newcommand{\Set}{\ensuremath{A}}
\newcommand{\TupleSet}{\ensuremath{T}}
\newcommand{\SetNew}{\ensuremath{\Set_\DomNew}\xspace}
\newcommand{\SetOld}{\ensuremath{\Set_\DomOld}\xspace}
\newcommand{\SetAll}{\ensuremath{\Set_\DomAll}\xspace}
\newcommand{\SetFact}{\ensuremath{\Set_\mathrm{f}}\xspace}
\newcommand{\SetDelta}{\ensuremath{\Set_\Delta}}
\newcommand{\TupleAll}{\ensuremath{\TupleSet_\DomAll}\xspace}
\newcommand{\TupleFact}{\ensuremath{\TupleSet_\mathrm{f}}\xspace}
\newcommand{\AggrAll}{\ensuremath{I}\xspace}
\newcommand{\AggrRec}{\ensuremath{I_\mathrm{r}}\xspace}
\newcommand{\Prog}{\ensuremath{P}\xspace}
\newcommand{\GroundProg}{\ensuremath{P_\mathrm{g}}\xspace}

\newcommand{\RecPred}{\ensuremath{A_\mathrm{r}}\xspace}

\DontPrintSemicolon
\SetKwProg{Fn}{\KwSty{function}}{}{end}
\SetKw{In}{in}
\SetKw{Where}{where}
\SetKw{With}{with}
\SetKw{Let}{let}
\SetKw{And}{and}
\SetKw{True}{true}
\SetKw{False}{false}
\SetKw{Or}{or}
\SetKw{Exists}{exists}
\SetKw{Not}{not}
\SetKw{Bot}{\ensuremath{\bot}}
\SetKwFunction{GroundProgram}{Ground}
\SetKwFunction{GroundRule}{GroundRule}
\SetKwFunction{AnalyzeProgram}{Analyze}
\SetKwFunction{PrepareComponent}{Prepare}
\SetKwFunction{PropagateAggregates}{Propagate}
\SetKwFunction{RewriteProgram}{Rewrite}
\SetKwFunction{AssembleAggregates}{Assemble}

\newcommand{\Matches}{\mathrm{matches}}
\newcommand{\Order}{\mathrm{order}}

\newcommand{\lparse}{\texttt{lparse}\xspace}
\newcommand{\gringo}{\texttt{gringo}\xspace}
\newcommand{\dlv}{\texttt{dlv}\xspace}
\newcommand{\gidl}{\texttt{gidl}\xspace}

\newfloatcommand{btabbox}{table}[][\FBwidth]
\newlength\hiwidth
\newlength\hiheight
\newlength\hidepth
\newlength\hitotal
\newcommand\HiLi{\leavevmode\rlap{\hbox to \hsize{\color{black!10}\leaders\hrule height \hiheight depth \hidepth\hfill}}}
\newcommand\HiWo[1]{\settowidth\hiwidth{#1}\leavevmode\rlap{\hbox{\color{black!10}\rule[-\hidepth]{\hiwidth}{\hitotal}}}#1}

\title{Grounding Recursive Aggregates: Preliminary Report}
\author{Martin Gebser\inst{1,3} \and Roland Kaminski\inst{3} \and Torsten Schaub\inst{2,3}\thanks{Affiliated with Simon Fraser University, Canada, and IIIS Griffith University, Australia.}} 
\institute{Aalto University, HIIT \and INRIA Rennes \and University of Potsdam}

\begin{document}

\maketitle

\begin{abstract}
Problem solving in Answer Set Programming consists of two steps,
a first grounding phase, systematically replacing all variables by terms,
and a second solving phase computing the stable models of the obtained ground program.
An intricate part of both phases is the treatment of aggregates,
which are popular language constructs that allow for expressing properties over sets.
In this paper, we elaborate upon the treatment of aggregates during grounding in \gringo~series 4.
Consequently, our approach is applicable to grounding based on semi-naive database evaluation techniques.
In particular, we provide a series of algorithms detailing the treatment of recursive aggregates
and illustrate this by a running example.
\end{abstract}


\section{Introduction}
\label{sec:introduction}

Modern grounders like (the one in) \texttt{dlv}~\cite{falepe12a} or \gringo~\cite{gekakosc11a} are based on semi-naive database evaluation techniques~\cite{ullman88a,abhuvi95a}
for avoiding duplicate work during grounding.
Grounding is seen as an iterative bottom-up process guided by the successive expansion of a program's Herbrand base, 
that is, the set of variable-free atoms constructible from the signature of the program at hand.
During this process,
a ground rule is only produced if its positive body atoms belong to the current Herbrand base,
in which case its head atom is added to the current Herbrand base.
The basic idea of semi-naive database evaluation is to focus this process on the new atoms generated at each iteration in order to avoid reproducing the same ground rules.
This idea is based on the observation that the production of a new ground rule relies on the existence of an atom having been new at the previous iteration.
Accordingly, a ground rule is only produced if its positive body contains at least one atom produced at the last iteration.

In what follows,
we show how a grounding framework relying upon semi-naive database evaluation techniques can be extended to incorporate recursive aggregates.
An example of such an aggregate is shown in Table~\ref{tab:company:encoding},
giving an encoding of the \emph{Company Controls Problem}~\cite{mupira90a}:
A company \(X\) controls a company \(Y\), 
if \(X\) directly or indirectly controls more than 50\% of the shares of \(Y\).
\begin{table}[ht]
\centering
\(\begin{aligned}
  \Pred{controls}(X,Y) \leftarrow {} & 
    \!\begin{aligned}[t]
      \Sum^+ \{ & S : \Pred{owns}(X,Y,S); \\
      & S,Z: \Pred{controls}(X,Z), \Pred{owns}(Z,Y,S) \} > \overline{50}
    \end{aligned} \\
  {} \wedge {} & \Pred{company}(X) \wedge \Pred{company}(Y) \wedge X \neq Y
\end{aligned}\)
\caption{Company Controls Encoding\label{tab:company:encoding}}
\end{table}
The aggregate \(\Sum^+\) implements summation over positive integers.
Notably, it takes part in the recursive definition of \Pred{controls}/2 in Table~\ref{tab:company:encoding}.
A corresponding problem instance is given in Table~\ref{tab:company:instance}.
\begin{table}[ht]
{\centering
\(\begin{aligned}
\Pred{company}(c_1)&. & \Pred{company}(c_2) & . & \Pred{company}(c_3) & . & \Pred{company}(c_4) & . \\
\Pred{owns}(c_1,c_2,\overline{60}) & . & \Pred{owns}(c_1,c_3,\overline{20}) & . & \Pred{owns}(c_2,c_3,\overline{35}) & . & \Pred{owns}(c_3,c_4,\overline{51}) & .
\end{aligned}\)\\}
\caption{Company Controls Instance\label{tab:company:instance}}
\end{table}
Note that a systematic instantiation of the four variables in Table~\ref{tab:company:encoding} with the eight constants in
Table~\ref{tab:company:instance} results in 64 ground rules.
However, taken together, the encoding and the instance are equivalent to the program in Table~\ref{tab:company:ground:naive},
which consists of four ground rules only.
\begin{table}[ht]
\centering
\(\begin{aligned}
  \Pred{controls}(c_1,c_2) \leftarrow {} & \Sum^+ \{ \overline{60} : \Pred{owns}(c_1,c_2,\overline{60}) \} > \overline{50} \\
  {} \wedge {} & \Pred{company}(c_1) \wedge \Pred{company}(c_2)\wedge c_1 \neq c_2 \\
  \Pred{controls}(c_3,c_4) \leftarrow {} & \Sum^+ \{ \overline{51} : \Pred{owns}(c_3,c_4,\overline{51}) \} > \overline{50} \\
  {} \wedge {} & \Pred{company}(c_3) \wedge \Pred{company}(c_4)\wedge c_3 \neq c_4 \\
  \Pred{controls}(c_1,c_3) \leftarrow {} & 
    \!\begin{aligned}[t]
      \Sum^+ \{ & \overline{20} : \Pred{owns}(c_1,c_3,\overline{20}); \\
                & \overline{35},c_2: \Pred{controls}(c_1,c_2), \Pred{owns}(c_2,c_3,\overline{35}) \} > \overline{50}
    \end{aligned} \\
  {} \wedge {} & \Pred{company}(c_1) \wedge \Pred{company}(c_3)\wedge c_1 \neq c_3 \\
  \Pred{controls}(c_1,c_4) \leftarrow {} & \Sum^+ \{ \overline{51},c_3: \Pred{controls}(c_1,c_3), \Pred{owns}(c_3,c_4,\overline{51}) \} > \overline{50} \\
  {} \wedge {} & \Pred{company}(c_1) \wedge \Pred{company}(c_4)\wedge c_1 \neq c_4
\end{aligned}\)
\caption{Relevant Grounding of Company Controls\label{tab:company:ground:naive}}
\end{table}
In fact,
all literals in Table~\ref{tab:company:ground:naive} can even be evaluated in view of the problem instance,
which moreover allows us to evaluate the aggregate atoms,
so that the grounding of the above company controls instance boils down to the four facts 
\(\Pred{controls}(c_1,c_2)\), \(\Pred{controls}(c_3,c_4)\), \(\Pred{controls}(c_1,c_3)\), and \(\Pred{controls}(c_1,c_4)\).

Accordingly,
the goal of this paper is to elaborate upon the efficient computation of the relevant grounding of programs with recursive aggregates.
Section~\ref{sec:background} starts with recalling the formal preliminaries from~\cite{gehakalisc15a}.
Section~\ref{sec:grounding} provides basic grounding algorithms (cf.~\cite{falepe12a}), paving the way 
for the more sophisticated algorithms addressing recursive aggregates in Section~\ref{sec:approach}.
We summarize our contribution and relate it to the state of the art in Section~\ref{sec:discussion}.
The developed approach is implemented in \gringo\ series~4.


\section{Formal Preliminaries}
\label{sec:background}

This section recalls the formal preliminaries regarding the syntax and semantics of \gringo's input language,
developed in~\cite{gehakalisc15a}.

\subsection{Syntax}

\paragraph{Alphabet.} 
We consider numerals, (symbolic) constants, variables, and aggregate names, along with the symbols
\begin{gather}
  \neq \qquad < \qquad > \qquad \leq \qquad \geq \label{symbol:comp}\footnotemark \\
  \bot \qquad \Naf \qquad \wedge \qquad \vee \qquad \leftarrow \label{symbol:connective} \\
  , \qquad ; \qquad : \qquad ( \qquad ) \qquad \{ \qquad \} \label{symbol:structure}
\end{gather}%
\footnotetext{Equality is not included here because it is treated specially in \gringo;
  a description is beyond the scope of this paper.}%
Numerals are strings of numbers optionally preceded with a minus symbol.
Constants are strings of letters, underscores, and numbers starting with a lowercase letter.
Variables are strings of letters, underscores, and numbers starting with an uppercase letter.%
\footnote{We use \_ to denote anonymous variables, i.e., each \_ stands for unique variable.}

\paragraph{Terms.}
Numerals, constants, and variables are terms.
Given a constant \(f\) and a term tuple \(\boldsymbol{t}\), 
\(f(\boldsymbol{t})\) is a term as well.
A variable-free term is said to be ground.

\paragraph{Interpretation of numerals and aggregates.}

A numeral \(\overline{n}\) corresponds to the integer \(n\).
There is a total order on all ground terms extending that on numerals, that is,
for any integers \(m\) and \(n\), \(\overline{m} \leq \overline{n}\) if \(m \leq n\).

For a ground term tuple \(\boldsymbol{t}\),
\(\Weight(\boldsymbol{t})\) is \(n\),
if the first element of \(\boldsymbol{t}\) is a numeral of form \(\overline{n}\), 
otherwise it is \(0\).

Each aggregate name \(\alpha\) is associated with a function \(\widehat{\alpha}\) 
from the set of sets of ground term tuples into the set of ground terms.
Given a set \(T\) of ground term tuples, we consider 
the aggregate names/functions defined by
\begin{itemize}
\item \(\widehat{\Sum}(T)=\overline{\Sigma_{\boldsymbol{t} \in T}\Weight(\boldsymbol{t})}\), if the subset of tuples with non-zero weights is finite, and \(\overline{0}\) otherwise;
\item \(\widehat{\Sum^+}(T)=\overline{\Sigma_{\boldsymbol{t} \in T,\Weight(\boldsymbol{t})>0}\Weight(\boldsymbol{t})}\), if the subset of tuples with positive weights is finite, and \(\infty\) otherwise.\footnote{\(\overline{n} < \infty\) holds for any numeral \(\overline{n}\).} 
\end{itemize}

\paragraph{Atoms and literals.}
Symbolic atoms have the form \(p(\boldsymbol{t})\) where \(p\) is a constant and \(\boldsymbol{t}\) is a term tuple.
Comparison atoms have form \(u_1 \prec u_2\) where \(u_1\) and \(u_2\) are terms.
We use atom \(\bot\) to denote a comparison atom that is false (e.g., \(0 > 0\)),
and use \(\top\) analogously.
Simple literals have form \(a\) or \(\Naf a\) where \(a\) is a symbolic or comparison atom.

Aggregate atoms have form 
\begin{equation}\label{eq:aggregate}
\alpha \{ \boldsymbol{t}_1 : \boldsymbol{L}_1; \dots; \boldsymbol{t}_n : \boldsymbol{L}_n \} \prec s  
\end{equation}
where
\begin{itemize}
\item \(n \geq 0\)
\item \(\alpha\) is an aggregate name
\item each \(\boldsymbol{t}_i\) is a term tuple
\item each \(\boldsymbol{L}_i\) is a tuple of simple literals
\item \(\prec\) is one of the symbols (\ref{symbol:comp})
\item \(s\) is a term, also called guard
\end{itemize}
Finally, literals have the form \(a\) or \(\Naf a\) where \(a\) is either a symbolic, comparison, or aggregate atom.%
\footnote{\gringo\ as well as its semantic underpinnings in~\cite{gehakalisc15a} also allow for double negated literals of form \(\Naf\Naf a\).}

\paragraph{Rules and programs.}
Rules are of form \(h \leftarrow l_1 \wedge \dots \wedge l_n \), where \(n \geq 0\), \(h\) is a symbolic atom, and each \(l_i\) is a literal.
A program is a finite set of rules.

\paragraph{Miscellaneous definitions.}
We use the following projection functions on rules.
\begin{itemize}
  \item \(\Head(r) = h\) in a rule \(r\) of the above form
  \item \(\Body(r) = \{ l_1, \dots, l_n \}\) in a rule \(r\) of the above form
  \item \(\Body^+(r) = \{ a\in \Body(r) \mid a \text{ is a symbolic atom}\}\)
  \item \(\Body^-(r) = \{ a \mid \Naf a \in \Body(r), a \text{ is a symbolic atom} \} \)
  \item \(\Body^\pm(r) = \Body^+(r) \cup \Body^-(r) \)
\end{itemize}

In the following, some body literals are marked.
The binary relation \(r \dagger l\) holds if the literal \(l \in \Body(r)\) of rule \(r\) is marked.
Marked body literals are indicated by \(l^\dagger\).

A substitution is a mapping from variables to (ground) terms.
We represent substitutions by sets of form \( \{x_1 \mapsto t_1, \dots, x_n \mapsto t_n\} \)
where \(n \geq 0\), each \(x_i\) is a variable, and each \(t_i\) is a ground term.
A substitution~\(\sigma\) of the above form applied to a literal~\(l\), written~\(l\sigma\), replaces all occurrences of variables \(x_i\) in \(l\) with corresponding terms~\(t_i\).

Moreover, we associate in what follows each occurrence of an aggregate in a logic program with a unique identifier.
We use \(\alpha_i\), \(\boldsymbol{x}_i\), and \(s_i\) to refer to the aggregate function, tuple of global variables, and guard 
of the aggregate occurrence identified by \(i\).
Furthermore, \((s_i)^{\boldsymbol{x}_i}_{\boldsymbol{g}}\) refers to the ground guard where the variables listed in tuple \(\boldsymbol{x}_i\) are replaced with the corresponding terms in tuple \(\boldsymbol{g}\) in \(s_i\).

An aggregate \(\alpha\) together with a relation \(\prec\) is monotone,
if for any sets \(T_1 \subseteq T_2\) of ground term tuples and ground term \(s\), we have that 
\(\alpha(T_1) \prec s\) implies \(\alpha(T_2) \prec s\).

\subsection{Semantics}
The semantics of programs rests upon a translation into (infinitary) propositional formulas along with their stable models~\cite{truszczynski12a}.
\paragraph{Ground simple literals} are mapped via \(\tau\) on propositional atoms as follows.
\begin{itemize}
  \item \(\tau(a) = a\) for (ground) symbolic atom \(a\)
  \item \(\tau(t_1 \prec t_2)\) is \(\top\), if the relation \(\prec\) holds between \(t_1\) and \(t_2\), and \(\bot\) otherwise
  \item \(\tau(\Naf a) = \neg \tau(a)\) for a literal \(\Naf a\)
\end{itemize}

\paragraph{Global variables.}
A variable is global
\begin{itemize}
  \item in a simple literal, if it occurs in the literal
  \item in an aggregate literal, if it occurs in the guard
  \item in a rule, if it is global in the head or a body literal 
\end{itemize}

\paragraph{Aggregate literals.}
The translation \(\tau\) extends to aggregate atoms~\(a\) as in~\eqref{eq:aggregate} as follows.
An instance of an aggregate element \(\boldsymbol{t} : \boldsymbol{L}\) is obtained by substituting all its variables with ground terms.
We let \(\tau \boldsymbol{L}\) stand for the conjunction of applications of \(\tau\) to the ground simple literals in \(\boldsymbol{L}\).

Let \(E\) be the set of all instances of aggregate elements in \(a\).
A set \(\Delta \subseteq E\) justifies \(a\),
if the relation \(\prec\) holds between \(\widehat{\alpha}\{ \boldsymbol{t} \mid (\boldsymbol{t} : \boldsymbol{L}) \in \Delta \}\) and the guard \(s\).
Then, \(\tau a\) is the conjunction of formulas 
\(
\bigwedge_{(\boldsymbol{t} : \boldsymbol{L}) \in \Delta} \tau \boldsymbol{L}
\rightarrow
\bigvee_{(\boldsymbol{t} : \boldsymbol{L}) \in E\setminus\Delta} \tau \boldsymbol{L}
\) 
for all sets \(\Delta \subseteq E\) that do not justify \(a\).

A negative aggregate literal \(\Naf a\) is treated analogous to a negative simple literal.

\paragraph{Rules and programs.}
An instance of a rule \(r\) is obtained by substituting all global variables with ground terms.
Then,
\(\tau r\) is the set of formulas 
\(
\tau l_1 \wedge \dots \wedge \tau l_n \rightarrow \tau h
\) 
for all instances
\(h \leftarrow l_1 \wedge \dots \wedge l_n \)
of rule \(r\),
and \(\tau P = \bigcup_{r\in P}\tau r\) for a program~$P$.

\paragraph{Stable models.}
The stable models of a logic program \(P\) are the stable models of the (infinitary) propositional formula \(\tau P\)~\cite{truszczynski12a}.

\subsection{Safety and Rule Dependency Graph}
A global variable is safe in a rule, if it is bound by a positive symbolic literal in the rule body.
A non-global variable is safe in an aggregate element,
if it is bound by a positive symbolic literal in the corresponding aggregate element.
A rule is safe, if all its variables are safe.
A program is safe, if all its rules are safe.
In what follows, we consider safe programs only.

The \emph{rule dependency graph} \(G=(V,E)\) of a (normal) logic program \(P\) is a directed graph such that
\(V = P\) and \(E = \{ (r_1,r_2) \in V \times V \mid l \in \Body^\pm(r_2), \Head(r_1) \text{ unifies } l \}\).%
\footnote{Unification assumes that variables in 
$r_1$ and $r_2$ are distinct, even if they have the same name.}
The positive rule dependency graph \(G^+\) is defined similarly but considers edges induced by positive literals only (\(l \in \Body^+(r_2)\)).


\section{Basic Grounding Algorithms}
\label{sec:grounding}

This section provides some basic algorithms underlying semi-naive evaluation based grounding (see also~\cite{falepe12a}).
All of them apply to normal logic programs and are thus independent of the treatment of recursive aggregates
described in the next section.

We illustrate the basic algorithms by means of a Hamiltonian cycle example%
\footnote{\url{https://en.wikipedia.org/wiki/Hamiltonian_path_problem}}
using the graph in Figure~\ref{fig:ham:instance:graph}.
This graph is represented by the problem instance in Table~\ref{tab:ham:instance}.
The actual problem encoding is given in~\eqref{eq:ham-rule1} to~\eqref{eq:ham-rule4} below.
The resulting Hamiltonian cycle is expressed through instances of predicate \Pred{path}/2;
a detailed discussion of such encodings can be found in~\cite{niemela99a,martru99a,gekakasc12a}.
\begin{figure}
\begin{floatrow}
\ffigbox{%
\begin{tikzpicture}[
    ->,
    x=2cm,
    y=2cm,
    node/.style={circle,minimum size=5mm,draw},
    start/.style={circle,double=white,minimum size=5mm,draw}]
  \node (c) [node] at (0,0) { c };
  \node (d) [node] at (1,0) { d };
  \node (a) [start]  at (0,1) { a };
  \node (b) [node] at (1,1) { b };
  \draw (a) -- (b);
  \draw (a) -- (c);
  \draw (c) -- (d);
  \draw (c) -- (a);
  \draw (b) -- (d);
  \draw (b) -- (c);
  \draw (d) -- (a);
\end{tikzpicture}
}{%
  \caption{Hamiltonian Cycle Instance (Graph)\label{fig:ham:instance:graph}}%
}
\btabbox{%
\(\begin{aligned}
\Pred{node}(a) & . & \Pred{node}(b) & . & \Pred{node}(c) & . \\
\Pred{node}(d) & . & \Pred{start}(a) & . & \Pred{edge}(a,b) & . \\
\Pred{edge}(a,c) & . & \Pred{edge}(b,c) & . & \Pred{edge}(b,d) & . \\
\Pred{edge}(c,a) & . & \Pred{edge}(c,d) & . & \Pred{edge}(d,a) & .
\end{aligned}\)
}{%
  \caption{Hamiltonian Cycle Instance\label{tab:ham:instance}}%
}
\end{floatrow}
\end{figure}

\begin{align}
\Pred{path}(X,Y) & \leftarrow \Pred{edge}(X,Y) \wedge \Naf \Pred{omit}(X,Y) \label{eq:ham-rule1} \\
\Pred{omit}(X,Y) & \leftarrow \Pred{edge}(X,Y) \wedge \Naf \Pred{path}(X,Y) \label{eq:ham-rule2} \\
& \leftarrow \Pred{path}(X,Y) \wedge \Pred{path}(X',Y) \wedge X < X' \\
& \leftarrow \Pred{path}(X,Y) \wedge \Pred{path}(X,Y') \wedge Y < Y' \\
\Pred{on\_path}(Y) & \leftarrow \Pred{path}(X,Y) \wedge \Pred{path}(Y,Z) \\
& \leftarrow \Pred{node}(X) \wedge \Naf \Pred{on\_path}(X) \\
\Pred{reach}(X) & \leftarrow \Pred{start}(X) \\
\Pred{reach}(Y) & \leftarrow \Pred{reach}(X) \wedge \Pred{path}(X,Y) \label{eq:ham-rule3} \\
& \leftarrow \Pred{node}(X) \wedge \Naf \Pred{reach}(X) \label{eq:ham-rule4}
\end{align}

\subsection{Analyzing Logic Programs}

The function~\AnalyzeProgram\ given in Algorithm~\ref{fun:analyze} 
takes a logic program~\Prog,
classifies occurrences of recursive symbolic atoms (\RecPred), and
groups rules into components suitable for successive grounding.
The classification of atoms can be used to apply on-the-fly simplifications in the following algorithms
(cf.\ Algorithm~\ref{fun:ground-rule}).

\AnalyzeProgram first determines the strongly connected components of the program's dependency graph (Lines \ref{fun:analyze:init-begin}-\ref{fun:analyze:init-end}).
This graph contains dependencies induced by both positive and negative literals.
The outer loop (Lines~\ref{fun:analyze:loop-outer-begin}-\ref{fun:analyze:loop-outer-end}) iterates over its components in topological order.%
\footnote{
A component \(C_1\) precedes 
\(C_2\) when there is an edge \((r_1,r_2)\) with \(r_1 \in C_1\) and \(r_2\in C_2\).}
Each component is then further refined 
in terms of its positive dependency graph (Lines~\ref{fun:analyze:refine-begin}-\ref{fun:analyze:refine-end}).

The set \RecPred of recursive symbolic atoms is determined in Line~\ref{fun:analyze:recpred}.
These are all body literals whose atom unifies with the head of a rule in the current or a following component.
Finally, the refined component together with its recursive atoms is appended to the list \(L\) in Line~\ref{fun:analyze:result}.
This list is the result of the algorithm returned in Line~\ref{fun:analyze:return}.

\begin{algorithm}[t]
  \caption{Analyze Logic Programs for Grounding\label{fun:analyze}}
  \Fn{\(\AnalyzeProgram(\Prog)\)}{
    \Let \(G\) be the dependency graph of \(\Prog\)\;\label{fun:analyze:init-begin}
    \noindent\phantom{\Let}\(S\) be the strongly connected components of \(G\)\;\label{fun:analyze:init-end}
    \(L \leftarrow []\)\;
    \ForEach{\(C\) \In \(S\)}{\label{fun:analyze:loop-outer-begin}
      \Let \(G^+\) be the positive dependency graph of \(C\)\;\label{fun:analyze:refine-begin}
      \noindent\phantom{\Let}\(S^+\) be the strongly connected components of \(G^+\)\;\label{fun:analyze:refine-end}
      \ForEach{\(C^+\) \In \(S^+\)}{
        \Let \(\RecPred = \{a \in \Body^\pm(r_2) \mid r_1 \in \Prog, r_2 \in C^+, \Head(r_1) \text{ unifies } a\}\)\;\label{fun:analyze:recpred}
        \((L,\Prog) \leftarrow (L + [(C^+,\RecPred)], \Prog \setminus C^+)\)\;\label{fun:analyze:result}
      }
    }\label{fun:analyze:loop-outer-end}
    \Return \(L\)\;\label{fun:analyze:return}
  }
\end{algorithm}

\begin{figure}[t]
\begin{tikzpicture}[
    ->,
    y=-8mm,
    node distance=5mm,
    node/.style={circle,minimum size=5mm,draw},
    start/.style={circle,double=bg,minimum size=5mm,draw}]
  \node (omit) at (0,0) [draw,anchor=west] {\(\Pred{omit}(X,Y) \leftarrow \Pred{edge}(X,Y) \wedge \Naf {\mathbf{path(X,Y)}}\)};
  \node [left=of omit] { Component\(_{1,1}\): };
  \node (path) at (0,1) [draw,anchor=west] {\(\Pred{path}(X,Y) \leftarrow \Pred{edge}(X,Y) \wedge \Naf \Pred{omit}(X,Y)\)};
  \node [left=of path] { Component\(_{1,2}\): };
  \node (inde) at (0,2) [draw,anchor=west] {\(\leftarrow \Pred{path}(X,Y) \wedge \Pred{path}(X',Y) \wedge X < X'\)};
  \node [left=of inde] { Component\(_{2,1}\): };
  \node (outd) at (0,3) [draw,anchor=west] {\(\leftarrow \Pred{path}(X,Y) \wedge \Pred{path}(X,Y') \wedge Y < Y'\)};
  \node [left=of outd] { Component\(_{3,1}\): };
  \node (onpa) at (0,4) [draw,anchor=west] {\(\Pred{on\_path}(Y) \leftarrow \Pred{path}(X,Y) \wedge \Pred{path}(Y,Z)\)};
  \node [left=of onpa] { Component\(_{4,1}\): };
  \node (conp) at (0,5) [draw,anchor=west] {\(\leftarrow \Pred{node}(X) \wedge \Naf \Pred{on\_path}(X)\)};
  \node [left=of conp] { Component\(_{5,1}\): };
  \node (star) at (0,6) [draw,anchor=west] {\(\Pred{reach}(X) \leftarrow \Pred{start}(X)\)};
  \node [left=of star] { Component\(_{6,1}\): };
  \node (reac) at (0,7) [draw,anchor=west] {\(\Pred{reach}(Y) \leftarrow {\mathbf{reach(X)}} \wedge \Pred{path}(X,Y)\)};
  \node [left=of reac] { Component\(_{7,1}\): };
  \node (crea) at (0,8) [draw,anchor=west] {\(\leftarrow \Pred{node}(X) \wedge \Naf \Pred{reach}(X)\)};
  \node [left=of crea] { Component\(_{8,1}\): };

  \coordinate (lomi) at ($(omit.west)-(3mm,0mm)$);
  \coordinate (romi) at ($(omit.east)+(3mm,0mm)$);
  \coordinate (upat) at ($(path.west)-(3mm,-1mm)$);
  \coordinate (dpat) at ($(path.west)-(3mm,+1mm)$);
  \coordinate (dsta) at ($(star.east)+(3mm,-1mm)$);
  \coordinate (usta) at ($(star.east)+(3mm,+1mm)$);
  \coordinate (rrea) at ($(reac.east)+(3mm,0)$);
  \coordinate (ronp) at ($(onpa.east)+(3mm,0)$);
  \coordinate (rdst) at ($(dsta)+(3mm,0)$);
  \draw [dashed] (omit) -- (lomi) -- (upat) -- (upat-|path.west);
  \draw [dashed] (path) -- (romi|-path) -- (romi) -- (omit);
  \draw (dpat -| path.west) -- (dpat) -- (dpat |- reac) -- (reac);
  \draw (dpat |- inde) -- (inde);
  \draw (dpat |- outd) -- (outd);
  \draw (dpat |- onpa) -- (onpa);
  \draw (star.east |- dsta) -- (dsta) -- (dsta |- reac.north);
  \draw [dashed] (star.east |- usta) -- (usta -| rrea) -- (rrea |- crea) -- (crea);
  \draw [-,dashed] (rrea) -- (reac);
  \draw [dashed] (onpa) -- (ronp) -- (ronp|-conp) -- (conp);
  \draw [-] (reac.north-|rdst) -- (rdst) -- (dsta);
  \fill (dpat |- inde) circle [radius=1pt]
        (dpat |- outd) circle [radius=1pt]
        (dpat |- onpa) circle [radius=1pt]
        (dsta) circle [radius=1pt]
        (rrea) circle [radius=1pt];
\end{tikzpicture}
\caption{Hamiltonian Cycle Dependency Graph\label{fig:ham-dep-graph}}
\end{figure}
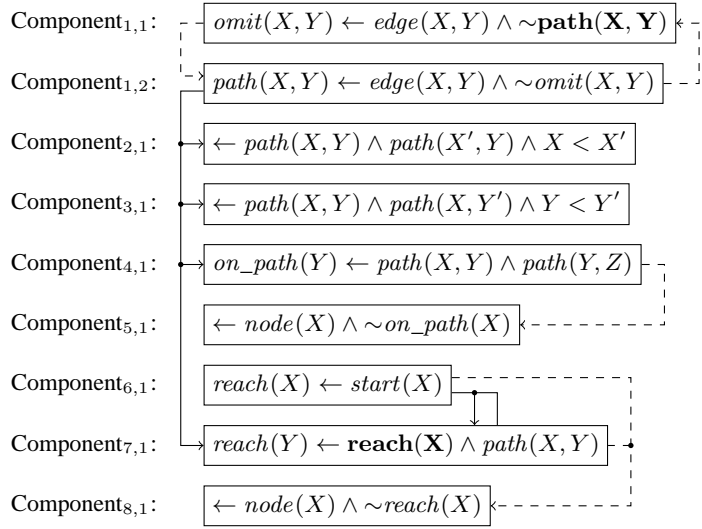
Figure~\ref{fig:ham-dep-graph} shows the dependency graph of the encoding given in \eqref{eq:ham-rule1} to~\eqref{eq:ham-rule4}.
Positive edges are depicted with solid lines, negative ones with dashed lines.
Recursive atoms are typeset in bold.
The negative edge from Component\(_{1,2}\) to Component\(_{1,1}\) is due to the fact that 
\(\Pred{path}(X,Y)\) in the negative body of~\eqref{eq:ham-rule2} unifies with 
\(\Pred{path}(X,Y)\) in the head of~\eqref{eq:ham-rule1}.
Furthermore, the occurrence of \(\Pred{path}(X,Y)\) in~\eqref{eq:ham-rule2} is recursive 
because it induces an edge from a later component in the topological ordering at hand.
In contrast to positive literals, the recursiveness of negative literals depends on the topological ordering.
For instance,
\(\Pred{omit}(X,Y)\) would be recursive in the topological order obtained by exchanging Component\(_{1,1}\) and Component\(_{1,2}\). 
%
Regarding Component\(_{7,1}\),
the occurrence of \(\Pred{reach}(X)\) in the body of \eqref{eq:ham-rule3} is recursive because it unifies with the head of the same rule.
Accordingly, it induces a self-loop in the dependency graph.

\subsection{Preparing Components for Grounding}

The function \PrepareComponent\ sets up the rules in a component~\(C\) for grounding w.r.t.\ its recursive atoms~\RecPred.
To this end,
it adds one of the subscripts \(\DomNew\), \(\DomOld\), or \(\DomAll\) to the predicate names of the atoms in the positive rule bodies of a given component.\footnote{%
The alphabet in Section~\ref{sec:background} does not allow for predicate names with subscripts.
During grounding, we temporarily extend this alphabet with such predicate names.}
These subscripts indicate \textit{new}, \emph{old}, and \emph{all} atoms belonging to the current materialization of the Herbrand base.
In turn, they are used in the course of semi-naive database evaluation to avoid duplicate work when grounding a component w.r.t.\ an expanding Herbrand base.

\begin{algorithm}[t]
  \caption{Prepare Components}
  \Fn{\(\PrepareComponent(C,\RecPred)\)}{
    \(L \leftarrow \emptyset\)\;
    \ForEach{\(r\) \In \(C\)}{\label{fun:prepare:loop-outer-begin}
      \(D \leftarrow \emptyset\)\;
      \Let \(S = \Body^+(r) \cap \RecPred\)\;\label{fun:prepare:pos-rec}
      \ForEach{\(p(\boldsymbol{x})\) \In \(S\)}{\label{fun:prepare:loop-inner-begin}
        \(L \leftarrow L \cup \left\{
          \textstyle\begin{aligned}
            \Head(r) \leftarrow {} & \textstyle\bigwedge_{q(\boldsymbol{y}) \in D} q_\DomOld(\boldsymbol{y}) \wedge p_\DomNew(\boldsymbol{x}) \\
            {} \wedge {}           & \textstyle\bigwedge_{q(\boldsymbol{y}) \in \Body^+(r) \setminus (D \cup \{p(\boldsymbol{x})\})} q_\DomAll(\boldsymbol{y}) \\
            {} \wedge {}           & \textstyle\bigwedge_{l \in \Body(r) \setminus \Body^+(r)} l
          \end{aligned}\right\}\)\;\label{fun:prepare:rule}
        \(D \leftarrow D \cup \{p(\boldsymbol{x})\}\)\;
      }\label{fun:prepare:loop-inner-end}
      \If{\(S = \emptyset\)}{\label{fun:prepare:check-nonrec}
        \(L \leftarrow L \cup
          \left\{
            \textstyle\begin{aligned}
              \Head(r) \leftarrow {} & \textstyle\bigwedge\nolimits_{ p(\boldsymbol{x}) \in \Body^+(r) } p_\DomNew(\boldsymbol{x}) \\
              {} \wedge {}           & \textstyle\bigwedge\nolimits_{ l \in \Body(r) \setminus \Body^+(r)} l
            \end{aligned}
          \right\}\)\;\label{fun:prepare:nonrec}
      }
    }\label{fun:prepare:loop-outer-end}
    \Return \(L\)\;
  }
\end{algorithm}

The loop in Lines \ref{fun:prepare:loop-outer-begin}-\ref{fun:prepare:loop-outer-end} iterates over the rules in the component at hand.
Each such rule \(r\) is expanded into a set of rules (loop in Lines \ref{fun:prepare:loop-inner-begin}-\ref{fun:prepare:loop-inner-end}) w.r.t.\ the recursive atoms in its body (Line~\ref{fun:prepare:pos-rec}).
In the first row of Line~\ref{fun:prepare:rule}, predicate names of recursive atoms already considered (set \(D\)) receive subscript~\DomOld,
and the predicate name of the recursive atom \(p(\boldsymbol{x})\) receives subscript \DomNew.
In the second row, the recursive atoms not yet considered as well as non-recursive atoms of the positive body receive subscript~\DomAll.
Finally, in the third row, the remaining body literals are kept unmodified.
If there are no recursive atoms (Line~\ref{fun:prepare:check-nonrec}),
then subscript \DomNew is added to all positive body elements (first row of Line~\ref{fun:prepare:nonrec}).
As in the case with recursive atoms,
the remaining body literals are kept unmodified (second row).

\begin{figure}[t]
\begin{tikzpicture}[
    ->,
    y=-8mm,
    node distance=5mm,
    node/.style={circle,minimum size=5mm,draw},
    start/.style={circle,double=bg,minimum size=5mm,draw}]
  \node (omit) at (0,0) [draw,anchor=west] {\(\Pred{omit}(X,Y) \leftarrow \Pred{edge}_\DomNew(X,Y) \wedge \Naf {\mathbf{path(X,Y)}}\)};
  \node [left=of omit] { Component\(_{1,1}\): };
  \node (path) at (0,1) [draw,anchor=west] {\(\Pred{path}(X,Y) \leftarrow \Pred{edge}_\DomNew(X,Y) \wedge \Naf \Pred{omit}(X,Y) \)};
  \node [left=of path] { Component\(_{1,2}\): };
  \node (inde) at (0,2) [draw,anchor=west] {\(\leftarrow \Pred{path}_\DomNew(X,Y) \wedge \Pred{path}_\DomNew(X',Y) \wedge X < X'\)};
  \node [left=of inde] { Component\(_{2,1}\): };
  \node (outd) at (0,3) [draw,anchor=west] {\(\leftarrow \Pred{path}_\DomNew(X,Y) \wedge \Pred{path}_\DomNew(X,Y') \wedge Y < Y'\)};
  \node [left=of outd] { Component\(_{3,1}\): };
  \node (onpa) at (0,4) [draw,anchor=west] {\(\Pred{on\_path}(Y) \leftarrow \Pred{path}_\DomNew(X,Y) \wedge \Pred{path}_\DomNew(Y,Z)\)};
  \node [left=of onpa] { Component\(_{4,1}\): };
  \node (conp) at (0,5) [draw,anchor=west] {\(\leftarrow \Pred{node}_\DomNew(X) \wedge \Naf \Pred{on\_path}(X)\)};
  \node [left=of conp] { Component\(_{5,1}\): };
  \node (star) at (0,6) [draw,anchor=west] {\(\Pred{reach}(X) \leftarrow \Pred{start}_\DomNew(X)\)};
  \node [left=of star] { Component\(_{6,1}\): };
  \node (reac) at (0,7) [draw,anchor=west] {\(\Pred{reach}(Y) \leftarrow {\mathbf{reach_\DomNew(X)}} \wedge \Pred{path}_\DomAll(X,Y)\)};
  \node [left=of reac] { Component\(_{7,1}\): };
  \node (crea) at (0,8) [draw,anchor=west] {\(\leftarrow \Pred{node}_\DomNew(X) \wedge \Naf \Pred{reach}(X)\)};
  \node [left=of crea] { Component\(_{8,1}\): };

  \node [right=of omit] (r1)   { \((r_1)\) };
  \node [at=(r1 |- path)] (r2) { \((r_2)\) };
  \node [at=(r2 |- inde)] (r3) { \((r_3)\) };
  \node [at=(r3 |- outd)] (r4) { \((r_4)\) };
  \node [at=(r4 |- onpa)] (r5) { \((r_5)\) };
  \node [at=(r5 |- conp)] (r6) { \((r_6)\) };
  \node [at=(r6 |- star)] (r7) { \((r_7)\) };
  \node [at=(r7 |- reac)] (r8) { \((r_8)\) };
  \node [at=(r8 |- crea)] (r9) { \((r_9)\) };

  \coordinate (lomi) at ($(omit.west)-(3mm,0mm)$);
  \coordinate (romi) at ($(omit.east)+(3mm,0mm)$);
  \coordinate (upat) at ($(path.west)-(3mm,-1mm)$);
  \coordinate (dpat) at ($(path.west)-(3mm,+1mm)$);
  \coordinate (dsta) at ($(star.east)+(3mm,-1mm)$);
  \coordinate (usta) at ($(star.east)+(3mm,+1mm)$);
  \coordinate (rrea) at ($(reac.east)+(3mm,0)$);
  \coordinate (ronp) at ($(onpa.east)+(3mm,0)$);
  \coordinate (rdst) at ($(dsta)+(3mm,0)$);
  \draw [dashed] (omit) -- (lomi) -- (upat) -- (upat-|path.west);
  \draw [dashed] (path) -- (romi|-path) -- (romi) -- (omit);
  \draw (dpat -| path.west) -- (dpat) -- (dpat |- reac) -- (reac);
  \draw (dpat |- inde) -- (inde);
  \draw (dpat |- outd) -- (outd);
  \draw (dpat |- onpa) -- (onpa);
  \draw (star.east |- dsta) -- (dsta) -- (dsta |- reac.north);
  \draw [dashed] (star.east |- usta) -- (usta -| rrea) -- (rrea |- crea) -- (crea);
  \draw [-,dashed] (rrea) -- (reac);
  \draw [dashed] (onpa) -- (ronp) -- (ronp|-conp) -- (conp);
  \draw [-] (reac.north-|rdst) -- (rdst) -- (dsta);
  \fill (dpat |- inde) circle [radius=1pt]
        (dpat |- outd) circle [radius=1pt]
        (dpat |- onpa) circle [radius=1pt]
        (dsta) circle [radius=1pt]
        (rrea) circle [radius=1pt];
\end{tikzpicture}
\caption{Prepared Hamiltonian Cycle Encoding\label{fig:ham-prepare}}
\end{figure}
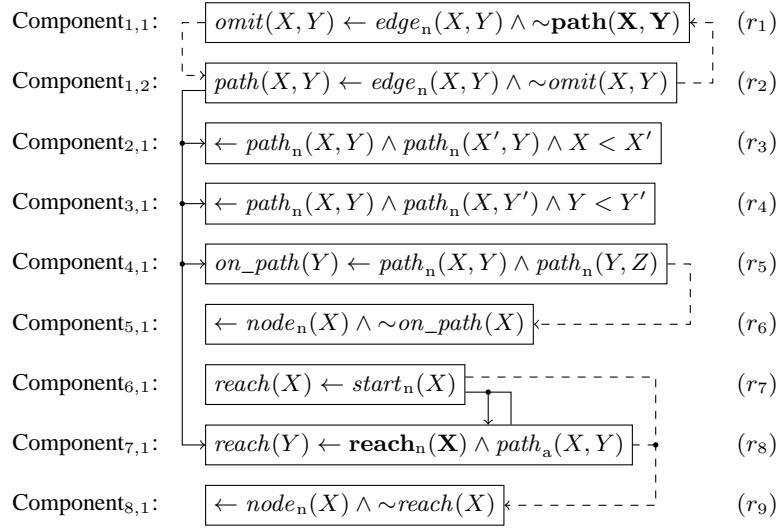
The result of preparing all components of the dependency graph in Figure~\ref{fig:ham-dep-graph} is given in Figure~\ref{fig:ham-prepare}.
All rules but \(r_8\) contain only non-recursive positive body literals,
which are adorned with subscript~\(\DomNew\).
Unlike this, 
the non-recursive positive body literal \(\Pred{path}(X,Y)\) in \(r_8\) is adorned with~\(\DomAll\),
while only the recursive one, \(\Pred{reach}(X)\), receives subscript~\(\DomNew\).
Since there is only one recursive body atom, only one rule is generated.

\subsection{Grounding Rules}

The rule grounding algorithm relies upon two auxiliary functions.
First,
function \(\Order\) returns a \emph{safe body order} of a rule body.%
\footnote{%
  The runtime of instantiation algorithms is sensitive to the chosen body order.
  In the context of ASP, heuristics for ordering body literals have been studied in \cite{lepesc01}.}
A safe body order of a body \(\{b_1, \dots, b_n\}\) is a tuple \((b_1, \dots, b_n)\) such that \(\{b_{1}, \dots, b_{i}\}\) is safe for each \(1\leq i\leq n\).
For example,
\((p(X),\Naf q(X))\) is a safe body order, while \((\Naf q(X),p(X))\) is not.
Second,
given a symbolic atom \(a\), a substitution \(\sigma\), and a set \(A\) of ground atoms,
function \(\Matches(a,\sigma,A)\) returns the set of \emph{matches} for \(a\) in \(A\) w.r.t. \(\sigma\).
A match is a \(\subseteq\)-minimal substitution \(\sigma'\) such that \(a\sigma' \in A\) and \(\sigma \subseteq \sigma'\).
For instance,
\(\Matches(p(X,Y),\{Y \mapsto a\},\linebreak[1]\{p(a,a),p(b,b),p(c,a)\})\) yields \(\{X\mapsto a,Y \mapsto a\}\) and \(\{X\mapsto c, Y \mapsto a\}\).

\begin{algorithm}[t]
  \caption{Grounding Rules\label{fun:ground-rule}}
  \Fn{\(\GroundRule(r,\RecPred,\SetNew,\SetOld,\SetAll,\SetFact)\)}{
    \Let \(r'\) be the original version of rule \(r\)\;
    \(G \leftarrow \emptyset\)\;
    \Fn{\(\GroundRule'(B,(b_1,\dots,b_n), \sigma)\)}{
      \uIf(\tcp*[f]{rule instance}){\(n=0\)}{
        \If{\(\Head(r)\sigma\not\in \SetFact\)}{
          \(G \leftarrow G \cup \{\Head(r)\sigma \leftarrow B\}\)\label{fun:ground-rule:append-rule}\;
          \(\SetFact \leftarrow \SetFact \cup \{ \Head(r)\sigma \mid B = \emptyset\}\)\label{fun:ground-rule:append-fact}\;
        }
      }
      \uElseIf(\tcp*[f]{positive literals}){\(b_1 = p_x(\boldsymbol{x})\) for \(x\in\{\DomOld, \DomNew, \DomAll\}\)}
      {
        \ForEach{\(\sigma' \in \Matches(p(\boldsymbol{x}),\sigma,A_x)\)}{
          \(B' \leftarrow B \cup \{p(\boldsymbol{x})\sigma' \mid r' \centernot\dagger p(\boldsymbol{x}),     p(\boldsymbol{x})\sigma'\notin\SetFact \}\)\label{fun:ground-rule:append-body-pos}\;
          \(\GroundRule'(B',(b_2,\dots,b_n),\sigma')\)\;
        }
      }
      \uElseIf(\tcp*[f]{negative literals}){\(b_1 = \Naf a\)}{
        \If{\(a\sigma \not\in \SetFact\)}{
          \(B' \leftarrow B \cup \{b_1\sigma \mid r' \centernot\dagger b_1,      a \in \RecPred~\Or~a\sigma\in \SetAll \}\)\label{fun:ground-rule:append-body-neg}\;
          \(\GroundRule'(B',(b_2,\dots,b_n),\sigma)\)\;
        }
      }
      \Else(\tcp*[f]{comparison atoms}){
        \If{\(b_1\sigma\) is true}{
          \(\GroundRule'(B,(b_2,\dots,b_n),\sigma)\)
        }
      }
    }
    \(\GroundRule'(\emptyset,\Order(\Body(r)),\emptyset)\)\label{fun:ground-rule:begin}\;
    \Return \((G,\SetFact)\)\;
  }
\end{algorithm}

With both functions at hand, we assemble the basic algorithm \GroundRule for grounding individual rules in Algorithm~\ref{fun:ground-rule}.
Note that the original rule \(r'\) along with its marking \(\dagger\) can be ignored in our context;
they are only relevant when treating aggregates in Section~\ref{sec:approach}.%
\footnote{%
This is the rule before the application of function \PrepareComponent.
That is, the positive body literals of \(r'\) are free of subscripts \DomOld, \DomNew, and \DomAll.
The conditions \(r' \centernot\dagger p(\boldsymbol{x})\) and \(r' \centernot\dagger b_1\) in Line~\ref{fun:ground-rule:append-body-pos} or~\ref{fun:ground-rule:append-body-neg}, respectively, are tautological in this section.}
The algorithm proceeds along the safe body order determined in Line~\ref{fun:ground-rule:begin}.
If no body literals remain,
a ground rule is generated in Line~\ref{fun:ground-rule:append-rule} provided that its head is not among the established facts.
Moreover, if the rule in focus has become a fact, its head is added to \SetFact in Line~\ref{fun:ground-rule:append-fact}.
The remainder constitutes a case analysis upon the type of the left-most body literal.
If \(b_1\) is a positive body literal, an instance of \(b_1\) is added in turn to the (partial) ground body \(B\) for each match of \(b_1\).
However, this is only done if the literal is not marked
and the instance does not yet belong to the established facts.
If \(b_1\) is a negative body literal, the instance obtained by applying the current substitution is added to the ground body.
Again, this is only done if the literal is not marked,
and the literal is recursive or there is already a derivation for it.
Substitutions where the instance is a fact are skipped altogether. 
Finally, comparison literals are directly evaluated and rule instantiation is only pursued if the test was successful.

\begin{figure}[t]
\centering
\begin{tikzpicture}[y=.7cm,x=2.2cm]
  \node (left) at (0,1) {\(\Pred{edge}_\DomNew(X,Y)\)};
  \node at (1,1) {\(\Naf \Pred{path}(X,Y)\)};
  \node at (2,1) {\(\Pred{omit}(X,Y)\)};
  \node[anchor=west] at (2.6,1) {\parbox{44mm}{
    \(\RecPred = \{ \Pred{path}(X,Y) \}\)
  }};

  \node at (0,0) (m1) {\(\Pred{edge}(a,b)\)};
  \node at (0,-1)(m2) {\(\Pred{edge}(a,c)\)};
  \node at (0,-2)(m3) {\(\Pred{edge}(b,c)\)};
  \node at (0,-3)(m4) {\(\Pred{edge}(b,d)\)};
  \node at (0,-4)(m5) {\(\Pred{edge}(c,a)\)};
  \node at (0,-5)(m6) {\(\Pred{edge}(c,d)\)};
  \node at (0,-6)(m7) {\(\Pred{edge}(d,a)\)};
  \draw (m1) -- (m2);
  \draw (m2) -- (m3);
  \draw (m3) -- (m4);
  \draw (m4) -- (m5);
  \draw (m5) -- (m6);
  \draw (m6) -- (m7);

  \node at (1,0) (m11) {\(\underline{\Naf \Pred{path}(a,b)}\)};
  \node at (1,-1)(m21) {\(\underline{\Naf \Pred{path}(a,c)}\)};
  \node at (1,-2)(m31) {\(\underline{\Naf \Pred{path}(b,c)}\)};
  \node at (1,-3)(m41) {\(\underline{\Naf \Pred{path}(b,d)}\)};
  \node at (1,-4)(m51) {\(\underline{\Naf \Pred{path}(c,a)}\)};
  \node at (1,-5)(m61) {\(\underline{\Naf \Pred{path}(c,d)}\)};
  \node at (1,-6)(m71) {\(\underline{\Naf \Pred{path}(d,a)}\)};
  \draw[->] (m1) -- (m11);
  \draw[->] (m2) -- (m21);
  \draw[->] (m3) -- (m31);
  \draw[->] (m4) -- (m41);
  \draw[->] (m5) -- (m51);
  \draw[->] (m6) -- (m61);
  \draw[->] (m7) -- (m71);

  \node at (2,0) (m111) {\(\Pred{omit}(a,b)\)};
  \node at (2,-1)(m211) {\(\Pred{omit}(a,c)\)};
  \node at (2,-2)(m311) {\(\Pred{omit}(b,c)\)};
  \node at (2,-3)(m411) {\(\Pred{omit}(b,d)\)};
  \node at (2,-4)(m511) {\(\Pred{omit}(c,a)\)};
  \node at (2,-5)(m611) {\(\Pred{omit}(c,d)\)};
  \node at (2,-6)(m711) {\(\Pred{omit}(d,a)\)};
  \draw[->] (m11) -- (m111);
  \draw[->] (m21) -- (m211);
  \draw[->] (m31) -- (m311);
  \draw[->] (m41) -- (m411);
  \draw[->] (m51) -- (m511);
  \draw[->] (m61) -- (m611);
  \draw[->] (m71) -- (m711);
  \node [fit=(m1)(m11)(m111), inner sep=0] (top) {};
  \node [anchor=north west] at (2.6,0 |- top.north) (right) {\parbox{44mm}{
    \(\SetFact = \left\{\!\begin{aligned}
        & \Pred{edge}(a,b), \Pred{edge}(a,c), \\
        & \Pred{edge}(b,c), \Pred{edge}(b,d), \\
        & \Pred{edge}(c,a), \Pred{edge}(c,d), \\
        & \Pred{edge}(d,a), \dots
      \end{aligned}\right\}\)\\
    \(\SetOld  = \emptyset\)\\
    \(\SetNew  = \SetFact\)\\
    \(\SetAll  = \SetFact\)
  }};
  \draw (0,0.5 -| left.west) -- (0,0.5 -| right.east);
\end{tikzpicture}
\caption{Call to \(\texttt{GroundRule}(r_1,\RecPred,\SetNew,\SetOld,\SetAll,\SetFact)\)\label{fig:ground-r1}}
\end{figure}
For illustration,
we trace in Figure~\ref{fig:ground-r1} the application of \GroundRule to rule~\(r_1\), viz.\
`\(\Pred{omit}(X,Y) \leftarrow \Pred{edge}_\DomNew(X,Y) \wedge \Naf \Pred{path}(X,Y)\)', from Figure~\ref{fig:ham-prepare}.
Figure~\ref{fig:ground-r1} also gives the contents of the respective sets of atoms (when tackling the very first component in Figure~\ref{fig:ham-prepare}).
The header in Figure~\ref{fig:ground-r1} contains the ordered body followed by the rule head.
Starting with the first positive body literal \(\Pred{edge}_\DomNew(X,Y)\) results in eight distinct matches of \(\Pred{edge}(X,Y)\) in \(A_\DomNew\).
The atoms resulting from a set of matches are connected with undirected edges in Figure~\ref{fig:ground-r1}.
Note that none of the instances of \(\Pred{edge}(X,Y)\) is  added to the (empty) body, since they are all found to be facts.
Looking at the trace for the first match,
we observe that \GroundRule' is next called with the empty body, the singleton \((\Naf \Pred{path}(X,Y))\), and substitution \(\{X\mapsto a,Y\mapsto b\}\).
Given that the atom \(\Pred{path}(a,b)\) is not a fact and recursive,
no simplifications apply, as indicated by underlining, and the instance is added to the body.
The following call with body \(\{\Naf {path(a,b)}\}\), the empty tuple, and the same substitution
     results in the ground rule `\(\Pred{omit}(a,b) \leftarrow \Naf {path(a,b)}\)'.
Analogously, the other six   matches result in further instances of \(r_1\).

\subsection{Grounding Logic Programs}

The above functions are put together in Algorithm~\ref{fun:ground} for grounding entire (normal) logic programs.
The function \GroundProgram takes a partition of a program into genuine rules \Prog and atoms \SetFact stemming from facts, 
and returns (upon termination) a set of ground instantiated rules \GroundProg.
The latter is incrementally constructed by following the topological order of components determined by \AnalyzeProgram.
Then, in turn, each adorned rule in the prepared component is instantiated via \GroundRule.
The loop in Lines~\ref{fun:ground:loop-bottomup-begin}-\ref{fun:ground:loop-bottomup-end} is executed only once whenever the component is free of recursive positive body literals,
and otherwise until no new (head) atoms are forthcoming.
This is accomplished by manipulating the following sets of atoms:
\begin{itemize}
  \item \(\SetAll\) the set of all relevant atoms up to the current grounding step,
  \item \(\SetNew \subseteq \SetAll\) the set of atoms atoms newly instantiated in the previous grounding step,
  \item \(\SetOld = \SetAll \setminus \SetNew\) the set of atoms that are not new w.r.t. the previous step,
  \item \(\SetDelta\) the set of atoms resulting from the current grounding step, and
  \item \(\SetFact \subseteq \SetAll \cup \SetDelta\) the set of atoms having a corresponding fact in \GroundProg.
\end{itemize}
The set \SetAll comprises the relevant Herbrand base when the algorithm terminates.
\begin{algorithm}[t]
  \caption{Grounding Logic Programs\label{fun:ground}}
  \Fn{\(\GroundProgram(\Prog, \SetFact)\)}{
    \((\GroundProg,\SetAll) \leftarrow (\emptyset,\SetFact)\)\;
    \ForEach{\((C,\RecPred)\) \In \(\AnalyzeProgram(\Prog)\)}{
      \((\SetNew,\SetOld) \leftarrow (\SetAll,\emptyset)\)\;
      \Repeat{\(\SetNew=\emptyset\) \Or \(\{r \in C \mid \Body^+(r) \cap \RecPred \neq \emptyset\} = \emptyset\)}{\label{fun:ground:loop-bottomup-begin}
        \(\SetDelta \leftarrow \emptyset\)\;
        \ForEach{\(r\) \In \(\PrepareComponent(C,\RecPred)\)}{
          \((\GroundProg', \SetFact) \leftarrow \GroundRule(r,\RecPred,\SetNew,\SetOld,\SetAll,\SetFact)\)\;
          \((\SetDelta,\GroundProg) \leftarrow (\SetDelta\cup\{\Head(r_\mathrm{g}) \mid r_\mathrm{g}\in \GroundProg'\}, \GroundProg \cup \GroundProg')\)\;
        }
        \((\SetNew,\SetOld,\SetAll) \leftarrow (\SetDelta \setminus \SetAll, \SetAll, \SetDelta \cup \SetAll)\)\;
      }
    }\label{fun:ground:loop-bottomup-end}
    \Return \(\GroundProg\)\;
  }
\end{algorithm}

For illustration,
let us trace \GroundProgram in Figure~\ref{fig:ground-c71} when grounding the last but one component from Figure~\ref{fig:ham-prepare}.
To be more precise,
this deals with the prepared version of Component\(_{7,1}\) containing rule \(r_8\) only, viz.\
`\(\Pred{reach}(Y) \leftarrow {{reach_\DomNew(X)}} \wedge \Pred{path}_\DomAll(X,Y)\)'.
The recursive nature of this rule results in three iterations of the loop in Lines~\ref{fun:ground:loop-bottomup-begin}-\ref{fun:ground:loop-bottomup-end}.
Accordingly, we index the atom sets in Figure~\ref{fig:ground-c71} to reflect their state in the respective iteration.
Moreover, we only provide the parts of \SetOld, \SetNew, \SetAll, and \SetFact that are relevant to grounding \(r_8\).
Otherwise, conventions follow the ones in Figure~\ref{fig:ground-r1}.
\begin{figure}[t]
\begin{tikzpicture}[y=.7cm,x=2.2cm,
  label/.style={xshift=14mm,anchor=west}]
  \newcommand\FigLabel[2]{\node[label] at (4,#1) {\strut\smaller\textbf{#2}};}
  \FigLabel{1}{1}
  \FigLabel{0}{1.1}
  \node at (0,1) (left) {\(\Pred{reach}_\DomNew(X)\)};
  \node at (1,1) {\(\Pred{path}_\DomAll(X,Y)\)};
  \node at (2,1) {\(\Pred{reach}(Y)\)};
  \node [anchor=west] at (2.55,1) {\parbox{5cm}{\(\RecPred = \{ \Pred{reach}(X) \}\)}};
  \node at (0,0) (m1) {\(\Pred{reach}(a)\)};

  \node at (1,0)  (m11) {\(\underline{\Pred{path}(a,b)}\)};
  \node at (1,-1) (m12) {\(\underline{\Pred{path}(a,c)}\)};

  \draw (m11) -- (m12);
  \draw[->] (m1) -- (m11);

  \node at (2,0)  (m111) {\(\Pred{reach}(b)\)};
  \node at (2,-1) (m121) {\(\Pred{reach}(c)\)};
  \draw[->] (m11) -- (m111);
  \draw[->] (m12) -- (m121);

  \node [fit=(m1)(m11)(m111), inner sep=0] (top) {};
  \node [anchor=north west] at (2.55,0 |- top.north) (right) {\parbox{48mm}{
      \(\SetFact^1 = \{ \Pred{reach}(a), \dots \}\)\\
      \(\SetOld^1  = \emptyset\)\\
      \(\SetNew^1 = \left\{\!\begin{aligned}
          & \Pred{path}(a,b), \Pred{path}(a,c), \\
          & \Pred{path}(b,c), \Pred{path}(b,d), \\
          & \Pred{path}(c,a), \Pred{path}(c,d), \\
          & \Pred{path}(d,a), \Pred{reach}(a), \dots
        \end{aligned}\right\}\)\\
      \(\SetAll^1  = \SetNew^1\)
  }};
  \draw (0,0.5 -| left.west) -- (0,0.5 -| right.east);
\begin{scope}[yshift=-3.27cm]
  \FigLabel{0}{1.2}
  \node at (0,0)  (m1) {\(\underline{\Pred{reach}(b)}\)};
  \node at (0,-2) (m2) {\(\underline{\Pred{reach}(c)}\)};

  \node at (1,0)  (m11) {\(\underline{\Pred{path}(b,c)}\)};
  \node at (1,-1) (m12) {\(\underline{\Pred{path}(b,d)}\)};

  \node at (1,-2) (m21) {\(\underline{\Pred{path}(c,a)}\)};
  \node at (1,-3) (m22) {\(\underline{\Pred{path}(c,d)}\)};

  \node at (2,0)  (m111) {\(\Pred{reach}(c)\)};
  \node at (2,-1) (m121) {\(\Pred{reach}(d)\)};

  \node at (2,-2) (m211) {\(\Pred{reach}(a)\)};
  \node at (2,-3) (m221) {\(\Pred{reach}(d)\)};

  \draw (m1) -- (m2);
  \draw (m11) -- (m12);
  \draw (m21) -- (m22);
  \draw[->] (m1) -- (m11);
  \draw[->] (m2) -- (m21);

  \draw[->] (m11) -- (m111);
  \draw[->] (m12) -- (m121);
  \draw[->] (m21) -- (m211);
  \draw[->] (m22) -- (m221);

  \node [fit=(m1)(m11)(m111), inner sep=0] (top) {};
  \node [anchor=north west] at (2.55,0 |- top.north) {\parbox{48mm}{%
     \(\SetFact^2 = \SetFact^1\)\\
     \(\SetOld^2 = \SetAll^1\)\\
     \(\SetNew^2 = \{ \Pred{reach}(b), \Pred{reach}(c) \}\)\\
     \(\SetAll^2 = \SetAll^1 \cup \SetNew^2\)
  }};
  \draw (0,0.5 -| left.west) -- (0,0.5 -| right.east);
\end{scope}
\begin{scope}[yshift=-6.07cm]
  \FigLabel{0}{1.3}
  \node at (0,0)  (m1) {\(\underline{\Pred{reach}(d)}\)};

  \node at (1,0)  (m11) {\(\underline{\Pred{path}(d,a)}\)};

  \node at (2,0) (m111) {\(\Pred{reach}(a)\)};

  \draw[->] (m1) -- (m11);

  \draw[->] (m11) -- (m111);

  \node [fit=(m1)(m11)(m111), inner sep=0] (top) {};
  \node [anchor=north west] at (2.55,0 |- top.north) {\parbox{48mm}{%
     \(\SetFact^3 = \SetFact^2\)\\
     \(\SetOld^3 = \SetAll^2\)\\
     \(\SetNew^3 = \{ \Pred{reach}(d) \}\)\\
     \(\SetAll^3 = \SetAll^2 \cup \SetNew^3\)
  }};
  \draw (0,0.5 -| left.west) -- (0,0.5 -| right.east);
\end{scope}
\end{tikzpicture}
\caption{Grounding Component\(_{7,1}\)\label{fig:ground-c71}}
\end{figure}
At the first iteration, 
the atom \(\Pred{reach}(a)\) (obtained from grounding Component\(_{6,1}\)) is used to obtain rule instances
\begin{align*}
  \Pred{reach}(b) & \leftarrow \Pred{path}(a,b)\\
  \Pred{reach}(c) & \leftarrow \Pred{path}(a,c)
\end{align*}
Note that \(\Pred{reach}(a)\) is removed from both rule bodies because it belongs to the established facts.
Moreover, this iteration yields the new atoms \(\Pred{reach}(b)\) and \(\Pred{reach}(c)\),
which are used in the next iteration to obtain the four rule instances
\begin{align*}
  \Pred{reach}(c) & \leftarrow \Pred{reach}(b) \wedge \Pred{path}(b,c)\\
  \Pred{reach}(d) & \leftarrow \Pred{reach}(b) \wedge \Pred{path}(b,d)\\
  \Pred{reach}(a) & \leftarrow \Pred{reach}(c) \wedge \Pred{path}(c,a)\\
  \Pred{reach}(d) & \leftarrow \Pred{reach}(c) \wedge \Pred{path}(c,d)
\end{align*}
Unlike above, no simplifications can be performed because no facts are involved.
The  iteration brings about a single new atom, \(\Pred{reach}(d)\),
which yields 
the rule instance
\begin{align*}
  \Pred{reach}(a) & \leftarrow \Pred{reach}(d) \wedge \Pred{path}(d,a)
\end{align*}
This iteration produces no new atoms and ends the instantiation of Component\(_{7,1}\).

The other components are grounded analogously but within a single iteration
due to their lack of recursive positive body literals.
This is enforced by the second stop criterion 
in Line~\ref{fun:ground:loop-bottomup-end} of Algorithm~\ref{fun:ground}.



\section{Grounding Recursive Aggregates}
\label{sec:approach}
 
Having laid the foundations of grounding normal logic programs,
we now continue with the treatment of recursive aggregates.
The idea is to translate aggregate atoms into normal logic programs,
roughly one rule per aggregate element,
and then to reuse the basic grounding machinery as much as possible.
In addition, some aggregate-specific propagation takes place.
At the end, the resulting aggregate instances are re-assembled from the corresponding rules.

\subsection{Rewriting Logic Programs with Aggregates}

The function~\(\RewriteProgram\) given in Algorithm~\ref{fun:rewrite} takes as input a logic program 
(possibly with recursive aggregates)
and rewrites it into a normal logic program with additional predicates
capturing aggregates and aggregate elements.%
\footnote{We assume that predicates \(\Aggr_i\) and \(\Accu_i\) are not used elsewhere in the program.}

\begin{algorithm}[t]
  \caption{Rewrite Logic Programs\label{fun:rewrite}}
  \Fn{\(\RewriteProgram(\Prog)\)}{
    \(Q \leftarrow \emptyset\)\;
    \tcp{in the loop below, \(\diamond\in\{\epsilon,{\Naf}\}\) stands for the sign of the aggregate literal}
    \ForEach{\(r\) \In \(\Prog\) \With \(a \in \Body(r)\), \(a = \diamond \alpha \{\boldsymbol{t}_1:\boldsymbol{L}_1; \dots; \boldsymbol{t}_n:\boldsymbol{L}_n\} \prec s\)}{
      \Let \(i\) be a unique identifier\;
      \noindent\phantom{\Let}\(\boldsymbol{x}\) be the global variables in \(a\)\;
      \noindent\phantom{\Let}\(B(\boldsymbol{L}) = \bigwedge_{ l \in \Body(r) \setminus \boldsymbol{L}, l \text{ is a simple literal}}l^\dagger\)\;
      replace occurrence \(a\) in \(\Prog\) with \(\diamond \Aggr_i(\boldsymbol{x})\)\;\label{fun:rewrite:replace}
      \(\!\begin{aligned}%
        Q \leftarrow Q & \cup \{ \Accu_i(\boldsymbol{x},\Neutral) \leftarrow \widehat\alpha(\emptyset) \prec s \wedge B(\emptyset) \}\\
                       & \cup \{ \Accu_i(\boldsymbol{x},\Tuple(\boldsymbol{t}_j)) \leftarrow \textstyle\bigwedge_{l \in \boldsymbol{L}_j}l \wedge B(\boldsymbol{L}_j) \mid 1 \leq j \leq n\} \\
                       & \cup \{ \Aggr_i(\boldsymbol{x}) \leftarrow \Accu_i(\boldsymbol{x},\_) \wedge \Bot \}
      \end{aligned}\)\;\label{fun:rewrite:elems}
    }
    \Return \(\Prog \cup Q\)
  }
\end{algorithm}

Each aggregate occurrence is replaced with an atom of form~\(\Aggr_i(\boldsymbol{x})\) in Line~\ref{fun:rewrite:replace},
where~\(i\) is a unique identifier associated with the aggregate occurrence and~\(\boldsymbol{x}\) are the global variables occurring in the aggregate.
The idea is that each atom over predicate~\(\Aggr_i\) in the grounding of the rewriting corresponds to a ground aggregate,
which is substituted for the atom in the final grounding.

To represent aggregate elements like $\boldsymbol{t}_j:\boldsymbol{L}_j$,
auxiliary rules defining atoms of form \(\Accu_i(\boldsymbol{x},t)\) are added in Line~\ref{fun:rewrite:elems},
where \(\boldsymbol{x}\) are the global variables as above and \(t\) is the tuple that is aggregated (or the special constant \(\Neutral\)).
Here, the idea is to inspect the grounding for rules with atoms over \(\Accu_i\) in the head.
If enough such atoms are accumulated to satisfy an aggregate,
then corresponding atoms over~\(\Aggr_i\) are added to the Herbrand base to further ground the program.

The first rule in Line~\ref{fun:rewrite:elems} handles the special case that the aggregate is satisfied for an empty set of tuples
(e.g., anti-monotone aggregates like~\(\Sum^+ \{ \boldsymbol{t}:\boldsymbol{L} \} \leq s\)).
Its body contains a comparison literal
that checks whether the empty aggregate is satisfied.
Furthermore, we have to make sure that the rule is safe so that it can be instantiated.
For this purpose, all simple literals of the rule in which the aggregate \(i\) occurs are added to the rule body (via function $B$).
Hence, if the original rule is safe, the auxiliary rule is also safe
because global variables are bound by positive symbolic literals only.
Furthermore, literals responsible for binding global variables are marked (via $\dagger$).

The second set of rules in Line~\ref{fun:rewrite:elems} is in charge of accumulating tuples of aggregate elements.
The rule body contains the literals of the condition of the aggregate element
as well as marked literals necessary for ensuring the rule's safety.
Remember that the resulting ground rules represent ground aggregate atoms, where
the marking is used to distinguish 
literals not belonging to the conditions of 
reconstituted aggregate elements.

Finally, one last rule is added in Line~\ref{fun:rewrite:elems}
for ensuring that the dependencies induced by the aggregate are kept intact.
Since this rule contains~\(\bot\), it never produces instances though.

\begin{table}[t]
{\centering
\(\begin{aligned}
  \Pred{controls}(X,Y)     & \leftarrow \Aggr_1(X,Y) \wedge B \\
  \Accu_1(X,Y,\Neutral)    & \leftarrow \overline{0} > \overline{50} \wedge B^\dagger \\
  \Accu_1(X,Y,\Tuple(S))   & \leftarrow \Pred{owns}(X,Y,S) \wedge B^\dagger \\
  \Accu_1(X,Y,\Tuple(S,Z)) & \leftarrow \Pred{controls}(X,Z) \wedge \Pred{owns}(Z,Y,S) \wedge B^\dagger \\
  \Aggr_1(X,Y)             & \leftarrow \Accu_1(X,Y,\_) \wedge \bot \\
\end{aligned}\)\\[8pt]}
where \(B = \Pred{company}(X) \wedge \Pred{company}(Y) \wedge X \neq Y\)
\caption{Rewritten Company Controls Encoding\label{tab:company:rewrite}}
\end{table}

The result of rewriting the company controls encoding from Table~\ref{tab:company:encoding} is given in Table~\ref{tab:company:rewrite}.
The global variables in the (single) aggregate are \(X\) and \(Y\), which occur first in all atoms over \(\Aggr_1\)/2 and \(\Accu_1\)/3.
Since the empty aggregate is not satisfied, the rule accumulating the \Neutral tuple never produces any instances (and could in principle be dropped from the rewriting).

\subsection{Analyzing and Preparing Logic Programs with Aggregates}

\begin{figure}[t]
  {\centering
  \begin{tikzpicture}[
    ->,
    y=-8mm,
    node distance=5mm,
    rule/.style={draw,anchor=north west}]
    \node (empty) at (0,0) [rule] {\(\Accu_1(X,Y,\Neutral)    \leftarrow \overline{0} > \overline{50} \wedge B^\dagger_\DomNew\)};
    \node [left=of empty] { Component\(_{1,1}\): };
    \node (strat) at (0,1) [rule] {\(\Accu_1(X,Y,\Tuple(S))   \leftarrow \Pred{owns}_\DomNew(X,Y,S) \wedge B^\dagger_\DomNew\)};
    \node [left=of strat] { Component\(_{2,1}\): };
    \node (close) at (0,2) [rule] {\(\Aggr_1(X,Y)             \leftarrow \mathbf{{accu_1}_\DomNew(X,Y,\_)} \wedge \bot\)};
    \node [left=of close] { Component\(_{3,1}\): };
    \node (rule)  at (0,3) [rule] {\(\Pred{controls}(X,Y)     \leftarrow \mathbf{{aggr_1}_\DomNew(X,Y)} \wedge B_\DomAll\)};
    \node (rec)   at (0,4) [rule]  {\(\!\begin{aligned}[t]\Accu_1(X,Y,\Tuple(S,Z)) \leftarrow {} & \mathbf{controls_\DomNew(X,Z)} \\ {} \wedge {} & \Pred{owns}_\DomAll(Z,Y,S) \wedge B^\dagger_\DomAll\end{aligned}\)};
    \coordinate (coord) at ($(close.west -| rec.west)-(.3cm,0)$);
    \coordinate (erule) at ($(rule.east)+(3mm,+1mm)$);
    \coordinate (frule) at ($(rule.east)+(3mm,-1mm)$);
    \draw [-] (strat) -- (strat -| coord);
    \draw (close) -| (erule) -- (rule.east |- erule);
    \draw (rule.east |- frule) -- (frule) -- (rec.north -| frule);
    \draw (rec) -| (coord) -- (close);
    \draw [-] (empty) -| (coord);
    \fill (coord) circle [radius=1pt];
    \fill (strat -| coord) circle [radius=1pt];
  \end{tikzpicture}\\[8pt]}
where \(B_x = \Pred{company}_x(X) \wedge \Pred{company}_x(Y) \wedge X \neq Y\)
  \caption{Dependency Graph for Company Controls Program\label{fig:company:analyze}}
\end{figure}
Figure~\ref{fig:company:analyze} captures the result of function~\AnalyzeProgram with \PrepareComponent called for each component of  the rewritten company controls encoding in Table~\ref{tab:company:rewrite}.
The rules in Component\(_{1,1}\) and Component\(_{2,1}\) depend on facts only, and thus both induce a singleton component.
Component\(_{3,1}\) contains the remaining rules.
The aggregate of the company controls encoding is recursive in this component
in view of the atom \(\Pred{controls}(X,Z)\) in its second aggregate element.
Note that not all aggregate elements are involved in this recursion, given that
    direct shares are accumulated via the rule in Component\(_{2,1}\).

\subsection{Propagating Aggregates}

\begin{algorithm}[t]
  \caption{Propagation of Aggregates\label{fun:propagate}}
  \Fn{\(\PropagateAggregates(I,r,\SetAll,\SetFact)\)}{
    \(\SetDelta \leftarrow \emptyset\)\;
    \ForEach{\(i,\boldsymbol{g}\) \Where \(i \in I\) \And \(\Accu_i(\boldsymbol{g},t) \in \SetAll\)}{\label{fun:propagate:loop-global-begin}
      \Let \(\TupleFact = \{ \boldsymbol{t} \mid \Accu_i(\boldsymbol{g},\Tuple(\boldsymbol{t})) \in \SetFact, \boldsymbol{t} \mbox{ is relevant for } \alpha_i  \}\)\label{fun:propagate:fact}\;
      \noindent\phantom{\Let}\(\TupleAll = \{ \boldsymbol{t} \mid \Accu_i(\boldsymbol{g},\Tuple(\boldsymbol{t})) \in \SetAll, \boldsymbol{t} \mbox{ is relevant for } \alpha_i \}\)\label{fun:propagate:unknown}\;
      \If{\Exists \( \TupleFact \subseteq \TupleSet \subseteq \TupleAll\) \Where \(\widehat{\alpha}_i(\TupleSet) \prec_i (s_i)^{\boldsymbol{x}_i}_{\boldsymbol{g}}\) is true}{\label{fun:propagate:check}
        \If{$($aggregate \(i\) is monotone \And \(\widehat{\alpha}_i(\TupleFact) \prec_i (s_i)^{\boldsymbol{x}_i}_{\boldsymbol{g}}\)$)$\label{fun:propagate:check-monotone} \\
          \noindent\phantom{If } \Or $($\Not \(r\) \And \(\TupleAll\setminus\TupleFact = \emptyset\)$)$\label{fun:propagate:check-stratified} }{
          \(\SetFact \leftarrow \SetFact \cup \{\Aggr_i(\boldsymbol{g})\}\)\label{fun:propagate:add-fact}\;
        }
        \(\SetDelta \leftarrow \SetDelta \cup \{\Aggr_i(\boldsymbol{g})\}\)\label{fun:propagate:add-delta}\;
      }
    }\label{fun:propagate:loop-global-end}
    \Return \((\SetDelta,\SetFact)\)\label{fun:propagate:result}\;
  }
\end{algorithm}
The function \PropagateAggregates inspects the partial grounding of an aggregate instance in view of its grounded aggregate elements.
To this end, it checks atoms over predicate \(\Accu_i\) obtained during grounding.
The loop in Lines~\ref{fun:propagate:loop-global-begin}-\ref{fun:propagate:loop-global-end} iterates over the given aggregate indices~\(I\)
and tuples of global variables stored in atoms over predicate \(\Accu_i\) appearing 
among the atoms in~\(A_\DomAll\).
Whenever there are enough tuples captured by such atoms to satisfy the corresponding aggregate,
\PropagateAggregates adds atoms over predicate \(\Aggr_i\) to \(\SetDelta\) for further instantiation.
While
Line~\ref{fun:propagate:fact} collects tuples that are necessarily accumulated by the aggregate function,
Line~\ref{fun:propagate:unknown} gathers tuples whose conditions 
can possibly hold.
Also note that
the relevance check skips tuples that do not change the result of an aggregate function.%
\footnote{For non-recursive aggregates, where the flag~$r$ is false,
Line~\ref{fun:propagate:check-stratified} checks whether \(\TupleAll\setminus\TupleFact\) is empty. 
  This is why only relevant tuples are gathered.}
For sum aggregates, this amounts to excluding zero-weight tuples
by stipulating \(\Weight(\boldsymbol{t}) \neq 0\).
Given these sets of tuples,
Line~\ref{fun:propagate:check} checks whether the aggregate can be satisfied using the tuples accumulated so far.
For sum aggregates, this can be tested by
adding  up the weights of factual tuples and, 
on the one hand, the negative weights to obtain a minimum, \(\mathit{min}\),
or 
                 the positive weights to obtain a maximum, \(\mathit{max}\).
Then, depending on the relation, the aggregate is satisfiable
\begin{itemize}
\item if \( \overline{\mathit{max}} \prec (s_i)^{\boldsymbol{x}_i}_{\boldsymbol{g}}\) is true for \({\prec} \in \{{\geq}, {>}\}\),
\item if \( \overline{\mathit{min}} \prec (s_i)^{\boldsymbol{x}_i}_{\boldsymbol{g}}\) is true for \({\prec} \in \{{\leq}, {<}\}\), or
\item if \( \overline{\mathit{min}} \prec (s_i)^{\boldsymbol{x}_i}_{\boldsymbol{g}}\) or \( \overline{\mathit{max}} \prec (s_i)^{\boldsymbol{x}_i}_{\boldsymbol{g}}\) is true for \({\prec} \in \{{\neq}\}\).
\end{itemize}
If the test in Line~\ref{fun:propagate:check} succeeds, 
the ground aggregate atom is added to the new atoms in Line~\ref{fun:propagate:add-delta}.
In addition, given a non-recursive or monotone aggregate, 
the corresponding ground aggregate atom is added to \SetFact whenever the aggregate is found to be true.
At this point, a non-recursive aggregate is true, if all its elements are facts  (Line~\ref{fun:propagate:check-stratified}),
and a monotone aggregate is true, if enough facts to satisfy the aggregate have been accumulated (Line~\ref{fun:propagate:check-monotone}).
Finally, the sets of new and factual atoms are returned in Line~\ref{fun:propagate:result}.

\subsection{Assembling Aggregates}

After \RewriteProgram has decomposed aggregate atoms into normal program rules,
\AssembleAggregates given in Algorithm~\ref{fun:assemble} reconstructs their grounded counterparts from the rewritten ground program.
That is,
all occurrences of atoms of form \(\Aggr_i(\boldsymbol{g})\) are replaced by their corresponding aggregates.
In doing so,
the aggregate elements are reconstructed from rules with head atoms \(\Accu_i(\boldsymbol{g},\Tuple(\boldsymbol{t}))\) in Line~\ref{fun:assemble:elements},
where 
an  element consists of the term tuple \(\boldsymbol{t}\) along with the condition expressed by the rule body.%
\footnote{Recall that marked literals, added for safety, are stripped off by \GroundRule in Algorithm~\ref{fun:ground-rule}.}
The actual replacement takes place in Line~\ref{fun:assemble:replace}, followed by the deletion of 
obsolete rules in
Line~\ref{fun:assemble:remove}.
Finally, the reconstructed ground program is returned in Line~\ref{fun:assemble:result}.
\begin{algorithm}[t]
  \caption{Assembling Aggregates\label{fun:assemble}}
  \Fn{\(\AssembleAggregates(\GroundProg)\)}{
    \ForEach{\(\Aggr_i(\boldsymbol{g})\) occurring in \(\GroundProg\)}{\label{fun:assemble:loop-begin}
      \tcp{below, \(\Body(r)\) is assumed to convert to a tuple of literals}
      \Let \(E = \{ \boldsymbol{t} : \Body(r) \mid r \in \GroundProg, \Head(r) = \Accu_i(\boldsymbol{g},\Tuple(\boldsymbol{t})) \}\)\label{fun:assemble:elements}\;
      replace all occurrences of \(\Aggr_i(\boldsymbol{g})\) in \(\GroundProg\) with \(\alpha_i  ( E  ) \prec_i (s_i)_{\boldsymbol{g}}^{\boldsymbol{x}_i}\)\label{fun:assemble:replace}\;
    }
    remove all rules with atoms over \(\Accu_i\)  in the head from \GroundProg\label{fun:assemble:remove}\;
    \Return \(\GroundProg\)\label{fun:assemble:result}\;
  }\label{fun:assemble:loop-end}
\end{algorithm}

\subsection{Grounding Logic Programs with Aggregates}

\begin{algorithm}[t]
  \caption{Grounding Logic Programs with Aggregates\label{fun:ground-aggregates}}
  \Fn{\(\GroundProgram(\Prog,\SetFact)\)}{
    \((\GroundProg,\SetAll) \leftarrow (\emptyset,\SetFact)\)\;
    \ForEach{\((C,\RecPred)\) \In \(\AnalyzeProgram(\HiWo{\ensuremath\RewriteProgram(\Prog)})\)}{\label{fun:ground-aggregates:rewrite}
      \HiLi\Let \(\AggrAll = \{ i \mid \Aggr_i \text{ occurs in a rule head in } C\}\)\label{fun:ground-aggregates:collect}\;
      \HiLi\noindent\phantom{\Let}\(\AggrRec = \{ i \mid r \in C, \Head(r) = \Accu_i(\boldsymbol{x},            t ), a \in \Body^+(r) \cap \RecPred, r \centernot\dagger a\}\)\label{fun:ground-aggregates:recursive}\;
      \((\SetNew,\SetOld) \leftarrow (\SetAll,\emptyset)\)\;
      \Repeat{\(\SetNew=\emptyset\) \Or \(\{r \in C \mid \Body^+(r) \cap \RecPred \neq \emptyset\} = \emptyset\)}{\label{fun:ground-aggregates:loop-bottomup-begin}
        \(\SetDelta \leftarrow \emptyset\)\;
        \ForEach{\(r\) \In \(\PrepareComponent(C,\RecPred)\)}{\label{fun:ground-aggregates:loop-groundstep-begin}
          \((\GroundProg', \SetFact) \leftarrow \GroundRule(r,\RecPred,\SetNew,\SetOld,\SetAll,\SetFact)\)\;
          \((\SetDelta,\GroundProg) \leftarrow (\SetDelta\cup\{\Head(r_\mathrm{g}) \mid r_\mathrm{g}\in \GroundProg'\}, \GroundProg \cup \GroundProg')\)\;
        }\label{fun:ground-aggregates:loop-groundstep-end}
        \HiLi\If{\(\SetDelta \subseteq \SetAll            \)}{\label{fun:ground-aggregates:propagate-stratified-begin}
          \HiLi\((\SetDelta,\SetFact) \leftarrow \PropagateAggregates(\AggrAll \setminus \AggrRec,\False,\SetAll,\SetFact)\)\label{fun:ground-aggregates:propagate-stratified-end}\;
        }
        \HiLi\If{\(\SetDelta \subseteq \SetAll            \)}{\label{fun:ground-aggregates:propagate-recursive-begin}
          \HiLi\((\SetDelta,\SetFact) \leftarrow \PropagateAggregates(\AggrAll \cap \AggrRec,\True,\SetAll,\SetFact)\)\label{fun:ground-aggregates:propagate-recursive-end}\;
        }
        \((\SetNew,\SetOld,\SetAll) \leftarrow (\SetDelta \setminus \SetAll, \SetAll, \SetDelta \cup \SetAll)\)\;
      }
    }\label{fun:ground-aggregates:loop-bottomup-end}
    \Return \(\HiWo{\ensuremath\AssembleAggregates(\GroundProg)}\)\label{fun:ground-aggregates:result}\;
  }
\end{algorithm}

Algorithm~\ref{fun:ground-aggregates} extends the \GroundProgram function in Algorithm~\ref{fun:ground}
to logic program with aggregates.
To this end, the extended \GroundProgram function uses algorithms \RewriteProgram, \PropagateAggregates, and \AssembleAggregates
from the previous subsections.
The changes in the algorithm are highlighted with a gray background,
while other parts are left untouched.

The first change is in Line~\ref{fun:ground-aggregates:rewrite},
where function \RewriteProgram is called to turn the logic program~\(P\) into a normal logic program before calling \AnalyzeProgram.
Then, Lines~\ref{fun:ground-aggregates:collect} and~\ref{fun:ground-aggregates:recursive}
are added,
just before the loop in charge of grounding each component.
Here, all aggregate indices that have to be propagated during the instantiation of a component are collected.
First, all indices for which a rule with \(\Aggr_i\) in the head appears in the component are gathered in~\AggrAll.
Remember that these rules do not contribute instances because \(\bot\) belongs to their bodies
(cf.\ third row in Line~\ref{fun:rewrite:elems} of Algorithm~\ref{fun:rewrite}).
Instead, \PropagateAggregates is adding atoms over \(\Aggr_i\) to \SetDelta.
Second, in Line \ref{fun:ground-aggregates:recursive}, aggregate indices associated with recursive aggregates are collected, where the (positive)
recursion involves some aggregate element
indicated by a rule with head atom \(\Accu_i(\boldsymbol{x},t)\),
yet without considering the auxiliary body part marked with \(\dagger\).

The collected indices are in Lines \ref{fun:ground-aggregates:propagate-stratified-begin}-\ref{fun:ground-aggregates:propagate-recursive-end} used
to propagate the corresponding aggregates in the current component.
Propagation of aggregates is triggered whenever no more new atoms are obtained in the grounding loop in Lines \ref{fun:ground-aggregates:loop-groundstep-begin}-\ref{fun:ground-aggregates:loop-groundstep-end}.
First, non-recursive aggregates \(\AggrAll \setminus \AggrRec\) are instantiated.
At this point, if there is at least one ground atom of form \(\Accu_i(\boldsymbol{g},t)\in\SetAll\),
all aggregate elements of the corresponding aggregate (uniquely determined by the terms~\(\boldsymbol{g}\) for global variables) have been gathered.
The aggregate  can thus be propagated, and function \PropagateAggregates can 
apply additional simplifications (cf.\ Line~\ref{fun:propagate:check-stratified} in Algorithm~\ref{fun:propagate}).
%
Afterwards, recursive aggregates \(\AggrAll \cap \AggrRec\) are propagated in Line~\ref{fun:ground-aggregates:propagate-recursive-end}.
In this case,
we cannot assume that all aggregate elements 
have already been accumulated,
and propagation can thus not use all of the simplifications applicable to non-recursive aggregates, where
the  distinction is implemented by setting the second argument of \PropagateAggregates to \True.
Finally, in Line~\ref{fun:ground-aggregates:result}, aggregates are reconstructed from the intermediate grounding by calling function \AssembleAggregates.

\begin{figure}[ht]
  \caption{Grounding Company Controls\label{fig:ground-company}}
  \begin{tikzpicture}[
    ->,
    x=2.0cm,
    y=-5mm,
    every node/.style={},
    c0/.style={},
    c1/.style={},
    c2/.style={xshift=-4mm},
    c3/.style={xshift=-8mm},
    c4/.style={xshift=-12mm},
    c5/.style={xshift=-9mm},
    label/.style={xshift=4mm,anchor=west}]
    \coordinate (left) at (-0.42,0);
    \coordinate (right) at (5.46,0);
    \coordinate (right2) at (5.25,0);

    \newcommand\FigLabel[2]{\node[label] at (5,#1) {\strut\smaller\textbf{#2}};}
    \begin{scope}
      \FigLabel{0}{1}
      \node[c0] at (0,0) {\(\overline{0} > \overline{50}\)};
      \node[c1] at (1,0) {\(c_\DomNew^\dagger(X)\)};
      \node[c2] at (2,0) {\(c_\DomNew^\dagger(Y)\)};
      \node[c3] at (3,0) {\(X \neq Y\)};
      \node[c5] at (5,0) {\(a_1(X,Y,n)\)};
      \draw[-] (0,0.5 -| left) -- (0,0.5 -| right); 
      \FigLabel{1}{1.1}
      \node[c0] (n1) at (0,1) {\(\times\)};
      \draw[-,double] (0,1.5 -| left) -- (0,1.5 -| right); 
    \end{scope}

    \begin{scope}[yshift=-10mm]
      \FigLabel{0}{2}
      \node[c0] at (0,0) {\(o_\DomNew(X,Y,S)\)};
      \node[c1] at (1,0) {\(c_\DomNew^\dagger(X)\)};
      \node[c2] at (2,0) {\(c_\DomNew^\dagger(Y)\)};
      \node[c3] at (3,0) {\(X \neq Y\)};
      \node[c5] at (5,0) {\(a_1(X,Y,t(S))\)};
      \draw[-] (0,0.5 -| left) -- (0,0.5 -| right); 
      \FigLabel{1}{2.1}
      \node[c0] (n1) at (0,1) {\(o(c_1,c_2,\overline{60})\)}; \node[c1] (n11) at (1,1) {\(c(c_1)\)}; \node[c2] (n111) at (2,1) {\(c(c_2)\)}; \node[c3] (n1111) at (3,1) {\(c_1 \neq c_2\)}; \node[c5] (n11111) at (5,1) {\(a_1(c_1,c_2,t(\overline{60}))\)};
      \node[c0] (n2) at (0,2) {\(o(c_1,c_3,\overline{20})\)}; \node[c1] (n21) at (1,2) {\(c(c_1)\)}; \node[c2] (n211) at (2,2) {\(c(c_3)\)}; \node[c3] (n2111) at (3,2) {\(c_1 \neq c_3\)}; \node[c5] (n21111) at (5,2) {\(a_1(c_1,c_3,t(\overline{20}))\)};
      \node[c0] (n3) at (0,3) {\(o(c_2,c_3,\overline{35})\)}; \node[c1] (n31) at (1,3) {\(c(c_2)\)}; \node[c2] (n311) at (2,3) {\(c(c_3)\)}; \node[c3] (n3111) at (3,3) {\(c_2 \neq c_3\)}; \node[c5] (n31111) at (5,3) {\(a_1(c_2,c_3,t(\overline{35}))\)};
      \node[c0] (n4) at (0,4) {\(o(c_3,c_4,\overline{51})\)}; \node[c1] (n41) at (1,4) {\(c(c_3)\)}; \node[c2] (n411) at (2,4) {\(c(c_4)\)}; \node[c3] (n4111) at (3,4) {\(c_3 \neq c_4\)}; \node[c5] (n41111) at (5,4) {\(a_1(c_3,c_4,t(\overline{51}))\)};
      \draw[-,double] (0,4.5 -| left) -- (0,4.5 -| right); 
      \draw (n1) -- (n11); \draw (n11) -- (n111); \draw (n111) -- (n1111); \draw (n1111) -- (n11111);
      \draw (n2) -- (n21); \draw (n21) -- (n211); \draw (n211) -- (n2111); \draw (n2111) -- (n21111);
      \draw (n3) -- (n31); \draw (n31) -- (n311); \draw (n311) -- (n3111); \draw (n3111) -- (n31111);
      \draw (n4) -- (n41); \draw (n41) -- (n411); \draw (n411) -- (n4111); \draw (n4111) -- (n41111);
      \draw[-] ($(n1.south)+(0,3pt)$) -- ($(n2.north)-(0,3pt)$);
      \draw[-] ($(n2.south)+(0,3pt)$) -- ($(n3.north)-(0,3pt)$);
      \draw[-] ($(n3.south)+(0,3pt)$) -- ($(n4.north)-(0,3pt)$);
    \end{scope}

    \begin{scope}[yshift=-35mm]
      \FigLabel{0}{3}
      \node[c0] at (0,0) {\(\bot\)};
      \node[c1] at (1,0) {\({a_1}_\DomNew(X,Y,\_)\)};
      \node[c5] at (5,0) {\(g_1(X,Y)\)};
      \draw[-,dashed] (0,0.5 -| left) -- (0,0.5 -| right2); 
      \node[c0] at (0,1) {\({g_1}_\DomNew(X,Y)\)};
      \node[c1] at (1,1) {\(c_\DomAll^\dagger(X)\)};
      \node[c2] at (2,1) {\(c_\DomAll^\dagger(Y)\)};
      \node[c3] at (3,1) {\(X \neq Y\)};
      \node[c5] at (5,1) {\(r(X,Y)\)};
      \draw[-,dashed] (0,1.5 -| left) -- (0,1.5 -| right2); 
      \node[c0] at (0,2)  {\(r_\DomNew(X,Z)\)};
      \node[c1] at (1,2)  {\(o_\DomAll(Z,Y,S)\)};
      \node[c2] at (2,2)  {\(c_\DomAll^\dagger(X)\)};
      \node[c3] at (3,2)  {\(c_\DomAll^\dagger(Y)\)};
      \node[c4] at (4,2)  {\(X \neq Y\)};
      \node[c5] at (5,2)  {\(a_1(X,Y,t(S,Z))\)};
      \draw[-] (0,2.5 -| left) -- (0,2.5 -| right); 
      \FigLabel{3}{3.1}
      \node[anchor=west] at (0,3 -| left) { \PropagateAggregates: \( \{g_1(c_1,c_2),g_1(c_3,c_4)\}\) };
      \draw[-] (0,3.5 -| left) -- (0,3.5 -| right); 
      \FigLabel{4}{3.2}
      \node[c0] at (0,4) {\(\times\)};
      \draw[-,dashed] (0,4.5 -| left) -- (0,4.5 -| right2); 
      \node[c0] (n1) at (0,5) {\(g_1(c_1,c_2)\)}; \node[c1] (n11) at (1,5) {\(c(c_1)\)}; \node[c2] (n111) at (2,5) {\(c(c_2)\)}; \node[c3] (n1111) at (3,5) {\(c_1 \neq c_2\)}; \node[c5] (n11111) at (5,5) {\(r(c_1,c_2)\)};
      \node[c0] (n2) at (0,6) {\(g_1(c_3,c_4)\)}; \node[c1] (n21) at (1,6) {\(c(c_3)\)}; \node[c2] (n211) at (2,6) {\(c(c_4)\)}; \node[c3] (n2111) at (3,6) {\(c_3 \neq c_4\)}; \node[c5] (n21111) at (5,6) {\(r(c_3,c_4)\)};
      \draw[-,dashed] (0,6.5 -| left) -- (0,6.5 -| right2); 
      \node[c0] at (0,7) {\(\times\)};
      \draw[-] (0,7.5 -| left) -- (0,7.5 -| right); 
      \draw (n1) -- (n11); \draw (n11) -- (n111); \draw (n111) -- (n1111); \draw (n1111) -- (n11111);
      \draw (n2) -- (n21); \draw (n21) -- (n211); \draw (n211) -- (n2111); \draw (n2111) -- (n21111);
      \draw[-] ($(n1.south)+(0,3pt)$) -- ($(n2.north)-(0,3pt)$);
      \FigLabel{8}{3.3}
      \node[c0] at (0,8) {\(\times\)};
      \draw[-,dashed] (0,8.5 -| left) -- (0,8.5 -| right2); 
      \node[c0] at (0,9) {\(\times\)};
      \draw[-,dashed] (0,9.5 -| left) -- (0,9.5 -| right2); 
      \node[c0] (n1) at (0,10) {\(r  (c_1,c_2)\)}; \node[c1] (n11) at (1,10) {\(o(c_2,c_3,\overline{35})\)}; \node[c2] (n111) at (2,10) {\(c(c_1)\)}; \node[c3] (n1111) at (3,10) {\(c(c_3)\)}; \node[c4] (n11111) at (4,10) {\(c_1 \neq c_3\)}; \node[c5] (n111111) at (5,10) {\(a_1(c_1,c_3,t(\overline{35},c_2))\)};
      \node[c0] (n2) at (0,11) {\(r  (c_3,c_4)\)}; \node[c1] (n21) at (1,11) {\(\times\)};
      \draw[-] (0,11.5 -| left) -- (0,11.5 -| right); 
      \draw (n1) -- (n11); \draw (n11) -- (n111); \draw (n111) -- (n1111); \draw (n1111) -- (n11111); \draw (n11111) -- (n111111);
      \draw (n2) -- (n21);
      \draw[-] ($(n1.south)+(0,3pt)$) -- ($(n2.north)-(0,3pt)$);
      \FigLabel{12}{3.4}
      \node[anchor=west] at (0,12 -| left) { \PropagateAggregates: \( \{g_1(c_1,c_3)\}\) };
      \draw[-] (0,12.5 -| left) -- (0,12.5 -| right); 
      \FigLabel{13}{3.5}
      \node[c0] at (0,13) {\(\times\)};
      \draw[-,dashed] (0,13.5 -| left) -- (0,13.5 -| right2); 
      \node[c0] (n1) at (0,14) {\(g_1(c_1,c_3)\)}; \node[c1] (n11) at (1,14) {\(c(c_1)\)}; \node[c2] (n111) at (2,14) {\(c(c_3)\)}; \node[c3] (n1111) at (3,14) {\(c_1 \neq c_3\)}; \node[c5] (n11111) at (5,14) {\(r(c_1,c_3)\)};
      \draw[-,dashed] (0,14.5 -| left) -- (0,14.5 -| right2); 
      \node[c0] at (0,15) {\(\times\)};
      \draw[-] (0,15.5 -| left) -- (0,15.5 -| right); 
      \draw (n1) -- (n11); \draw (n11) -- (n111); \draw (n111) -- (n1111); \draw (n1111) -- (n11111);
      \FigLabel{16}{3.6}
      \node[c0] at (0,16) {\(\times\)};
      \draw[-,dashed] (0,16.5 -| left) -- (0,16.5 -| right2); 
      \node[c0] at (0,17) {\(\times\)};
      \draw[-,dashed] (0,17.5 -| left) -- (0,17.5 -| right2); 
      \node[c0] (n1) at (0,18) {\(r  (c_1,c_3)\)}; \node[c1] (n11) at (1,18) {\(o(c_3,c_4,\overline{51})\)}; \node[c2] (n111) at (2,18) {\(c(c_1)\)}; \node[c3] (n1111) at (3,18) {\(c(c_4)\)}; \node[c4] (n11111) at (4,18) {\(c_1 \neq c_4\)}; \node[c5] (n111111) at (5,18) {\(a_1(c_1,c_4,t(\overline{51},c_3))\)};
      \draw[-] (0,18.5 -| left) -- (0,18.5 -| right); 
      \draw (n1) -- (n11); \draw (n11) -- (n111); \draw (n111) -- (n1111); \draw (n1111) -- (n11111); \draw (n11111) -- (n111111);
      \FigLabel{19}{3.7}
      \node[anchor=west] at (0,19 -| left) { \PropagateAggregates: \( \{g_1(c_1,c_4)\}\) };
      \draw[-] (0,19.5 -| left) -- (0,19.5 -| right); 
      \FigLabel{20}{3.8}
      \node[c0] at (0,20) {\(\times\)};
      \draw[-,dashed] (0,20.5 -| left) -- (0,20.5 -| right2); 
      \node[c0] (n1) at (0,21) {\(g_1(c_1,c_4)\)}; \node[c1] (n11) at (1,21) {\(c(c_1)\)}; \node[c2] (n111) at (2,21) {\(c(c_4)\)}; \node[c3] (n1111) at (3,21) {\(c_1 \neq c_4\)}; \node[c5] (n11111) at (5,21) {\(r(c_1,c_4)\)};
      \draw[-,dashed] (0,21.5 -| left) -- (0,21.5 -| right2); 
      \node[c0] at (0,22) {\(\times\)};
      \draw[-] (0,22.5 -| left) -- (0,22.5 -| right); 
      \draw (n1) -- (n11); \draw (n11) -- (n111); \draw (n111) -- (n1111); \draw (n1111) -- (n11111);
      \FigLabel{23}{3.9}
      \node[c0] at (0,23) {\(\times\)};
      \draw[-,dashed] (0,23.5 -| left) -- (0,23.5 -| right2); 
      \node[c0] at (0,24) {\(\times\)};
      \draw[-,dashed] (0,24.5 -| left) -- (0,24.5 -| right2); 
      \node[c0] (n1) at (0,25) {\(r(c_1,c_4)\)}; \node[c1] (n11) at (1,25) {\(\times\)};
      \draw (n1) -- (n11);
    \end{scope}
  \end{tikzpicture}
  \(a_1 = \Accu_1\),
  \(g_1 = \Aggr_1\),
  \(r = \Pred{controls}\),
  \(c = \Pred{company}\),
  \(o = \Pred{owns}\),
  \(n = \Pred{neutral}\),
  \(t = \Pred{tuple}\)
\end{figure}
Figure~\ref{fig:ground-company} traces the whole grounding process of the rewritten company controls encoding given in Figure~\ref{fig:company:analyze}.
      The grounding of each component is separated by   a horizontal double line, and
the instantiation         of a component is shown analogously to Figure~\ref{fig:ground-c71}.
Due to lack of space, 
we refrain from giving sets \RecPred, \SetNew, \SetOld, and \SetAll.
For each component, indicated by its number on the right,
the contained rules are listed first,
followed by grounding iterations for the component in focus.
Each iteration is separated by a solid line and indexed with an iteration number
on the right, where (instances of) the three rules in Component\(_{3,1}\) are separated by dashed lines.
The symbol \(\times\) indicates that a body literal does not match,
which corresponds to the case that the \GroundRule' function in Algorithm~\ref{fun:ground-rule} backtracks.

The grounding of Component\(_{1,1}\) produces no rule instances
because the comparison atom    \(\overline{0} > \overline{50}\) is false.
By putting this literal first in the safe body order,
the remaining rule body can be completely ignored.
Next, in the grounding of Component\(_{2,1}\), direct shares given by facts over \Pred{owns}/3 are accumulated,
where the obtained atoms over \(\Accu_1\)/3 are classified as facts as well.
Then, we trace the grounding of Component\(_{3,1}\).
In the first iteration,
none of the rules produces instances because there are no atoms over \Pred{controls}/2 and \(\Aggr_1\)/2 yet.
Hence, aggregate propagation is triggered,
resulting in aggregate atoms 
\(\Aggr_1(c_1,c_2)\) and \(\Aggr_1(c_3,c_4)\),
for which enough shares have been accumulated upon grounding Component\(_{2,1}\).
Note that, since the aggregate is monotone, both atoms are established as facts.
In the second iteration,
the newly obtained aggregate atoms are used to instantiate the second rule of the component,
leading to new atoms over \Pred{controls}/2.
Observe that, by putting      \(\Aggr_1(X,Y)\) first in the safe body order, \GroundRule can instantiate the rule without backtracking.
In the third iteration,
the newly obtained atoms over \Pred{controls}/2 yield 
\(\Accu_1(c_1,c_3,t(\overline{35},c_2))\) via an
instance of the third rule of the component,
which in turn leads to the aggregate atom 
\(\Aggr_1(c_1,c_3)\).
The following iterations
proceed in a similar fashion until no new tuples are accumulated
and the grounding loop terminates. 
Confined to the original predicate \Pred{controls}/2,
the instantiation generates four atoms,
\(\Pred{controls}(c_1,c_2)\), \(\Pred{controls}(c_3,c_4)\), \(\Pred{controls}(c_1,c_3)\), and \(\Pred{controls}(c_1,c_4)\),
all of which are produced   as facts.

Note that the utilized safe body order affects the amount of backtracking in rule instantiation~\cite{lepesc01}.
One particular strategy used in \gringo is to prefer recursive atoms with subscript~\DomNew when ordering a rule body.
As        seen in the grounding of Component\(_{3,1}\) above,
this helps to avoid backtracking upon generating new rule instances.
Furthermore, for the company controls encoding,
simplifications ensure that the program is evaluated to facts.
In general, this is guaranteed for programs with stratified negation and monotone aggregates only~\cite{alcafalepe11a}.

\section{Discussion}
\label{sec:discussion}

We presented an algorithmic framework for grounding logic programs based on semi-naive database evaluation techniques.
Our framework, which is implemented in \gringo~series~4,
constitutes the first approach capturing full-fledged aggregates under Ferraris' semantics~\cite{ferraris11a,gehakalisc15a}.
%
While semi-naive evaluation techniques trace back to the field of database systems~\cite{ullman88a,abhuvi95a},
their introduction to grounding in ASP was pioneered by the \dlv system~\cite{dlv03a},
laying out basic semi-naive grounding algorithms (cf.~\cite{falepe12a}) similar to those in Section~\ref{sec:grounding}.
Given this proximity, our grounding techniques for handling recursive aggregates
could be adopted within \dlv, 
which is so far restricted to stratified aggregates.
Other grounding approaches are pursued in \gidl~\cite{wimade10a}, \lparse~\cite{syrjanen01a}, and earlier versions of
\gringo~\cite{gescth07a,gekaosscth09a,gekakosc11a}.
The latter two also support recursive (convex) aggregates but are limited by the necessity 
to bind non-global variables 
by domain predicates,
given that programs have to be \(\omega\)-~\cite{syrjanen01a} or \(\lambda\)-restricted~\cite{gescth07a}, respectively.
Unlike this, our approach merely relies on the safety condition and no further restriction is imposed on the input language.

Regarding the implementation, our approach aims at reusing existing grounding techniques for (normal) logic programs.
To this end, programs with aggregates are rewritten, 
and conventional semi-naive evaluation is extended with a propagation step for aggregates.
Eventually, ground aggregates are reconstructed from the obtained rule instances in a post-processing step.
While the present paper considered \Sum and \(\Sum^+\) aggregates only,
our approach is applicable to any aggregate function.
In fact, \Count, \Min, and \Max aggregates are also supported in \gringo~series~4,
and it is easily amenable to further aggregates
(by extending the \PropagateAggregates function).


\paragraph{Acknowledgments} 
This work was funded by AoF (grant 251170), DFG (grants SCHA 550/8 and 550/9),
as well as DAAD and AoF (joint project 57071677\,/\,279121).




\end{document}